\documentclass[sigconf]{acmart}
\AtBeginDocument{%
  \providecommand\BibTeX{{%
    \normalfont B\kern-0.5em{\scshape i\kern-0.25em b}\kern-0.8em\TeX}}}

\usepackage{amsmath}
\usepackage{graphicx}
\usepackage{tikz,graphics,color,fullpage,float,epsf,caption,subcaption}

\begin{document}

\title{Are Word Embedding Methods Stable and Should We Care About It?}


\author{Angana Borah}

%\email{anganaborah9@gmail.com}
\affiliation{%
  \institution{National Institute of Technology Silchar}
  \city{Silchar}
  \state{Assam}
  \country{India}
  \postcode{788010}
}

\author{Manash Pratim Barman}

%\email{manashpratim2209@gmail.com}
\affiliation{%
  \institution{Indian Institute of Information Technology Guwahati}
  \city{Guwahati}
  \state{Assam}
  \country{India}
  \postcode{781015}
}

\author{Amit Awekar}

\email{awekar@iitg.ac.in}
\affiliation{%
  \institution{Indian Institute of Technology Guwahati}
  \city{Guwahati}
  \state{Assam}
  \country{India}
  \postcode{781039}
}

\renewcommand{\shortauthors}{Borah et al.}

\begin{abstract}
  A representation learning method is considered stable if it consistently generates similar representation of the given data across multiple runs. Word Embedding Methods (WEMs) are a class of representation learning methods that generate dense vector representation for each word in the given text data. The central idea of this paper is to explore the stability measurement of WEMs using intrinsic evaluation based on word similarity. We experiment with three popular WEMs: Word2Vec, GloVe, and fastText. For stability measurement, we investigate the effect of five parameters involved in training these models. We perform experiments using four real-world datasets from different domains: Wikipedia, News, Song lyrics, and European parliament proceedings. We also observe the effect of WEM stability on three downstream tasks: Clustering, POS tagging, and Fairness evaluation. Our experiments indicate that amongst the three WEMs, fastText is the most stable, followed by GloVe and Word2Vec.
\end{abstract}

%\begin{CCSXML}
%<ccs2012>
%  <concept_id>10010520.10010553.10010562</concept_id>
%  <concept_desc>Computer systems organization~Embedded systems</concept_desc>
%  <concept_significance>500</concept_significance>
%  <concept_id>10010520.10010575.10010755</concept_id>
%  <concept_desc>Computer systems organization~Redundancy</concept_desc>
%  <concept_significance>300</concept_significance>
%  <concept_id>10010520.10010553.10010554</concept_id>
%  <concept_desc>Computer systems %organization~Robotics</concept_desc>
%  <concept_significance>100</concept_significance>
%  <concept_id>10003033.10003083.10003095</concept_id>
%  <concept_desc>Networks~Network reliability</concept_desc>
%  <concept_significance>100</concept_significance>
%</ccs2012>
%\end{CCSXML}

%\ccsdesc[500]{Computer systems organization~Embedded systems}


\maketitle

\section{Introduction}
Word Embedding Methods (WEMs) are based on the idea that contextual information alone constitutes a feasible representation of linguistic terms. For each word in the vocabulary, these methods learn dense and low-dimensional vector representations, also referred as embeddings . These embeddings are observed to capture various semantic properties of words. In distributional semantics, vector space models are being used since the 1990s to capture contextual features. Some of the influential early models include the Latent Semantic Analysis \cite{deerwester1990indexing} and Latent Dirichlet Allocation \cite{blei2003latent}. But the popularization and utilization of word embeddings in natural language processing (NLP) tasks can be attributed to Mikolov and others who created Word2Vec \cite{NIPS2013_5021}. Two other popular WEMs are GloVe \cite{pennington2014glove} and fastText \cite{fasttext}.

The quick answer to the questions in the title of this paper is: WEMs are not stable and we should care about it. Given a text corpus $C$ and a WEM $A$, we can train multiple sets of word embeddings by varying different parameters related to $C$ and $A$. Each such set corresponds to an embedding space. Consider a word $w$ and two embedding spaces $E_1$ and $E_2$. The representation or embedding of $w$ in respective spaces be $E_1(w)$ and $E_2(w)$. We can evaluate the stability of representation of $w$ across $E_1$ and $E_2$ by comparing various properties of $E_1(w)$ and $E_2(w)$. The overall stability of WEM $A$ over corpus $C$ can be computed as an aggregate over the vocabulary of $C$. Several benchmarks  such as WordSim-353 \cite{finkelstein2002placing} and MEN \cite{bruni2014multimodal} exist for evaluating the quality of word embeddings. The former contains 353-word pairs while the latter contains 3000-word pairs. However, evaluating models using these tests gives only a partial representation of the model performance. In this work, rather than focusing on only a small subset of words, we focus on evaluating the stability of each word in the vocabulary.

Stability evaluation of WEMs has received a lot of attention in the recent past \cite{antoniak2018evaluating, dridi2018k, chugh, levy2015improving, wendlandt2018factors, lee2019understanding}. Such evaluation is crucial because word embeddings are widely used today for a variety of NLP tasks such as sentiment analysis \cite{tang2014learning, li2017learning}, named entity recognition \cite{lample2016neural, seok2016named} and part-of-speech (POS) tagging \cite{wendlandt2018factors, wang2015unified}. However, WEMs are sensitive to a number of factors including the size of the corpus, seeds for random number generations and other algorithmic parameters like number of vector dimensions, size of context window and number of training epochs. This leads to a question: what if the WEMs themselves are not stable? This could have implications for downstream tasks.

\begin{table*}[hbt]
\centering
%\resizebox{0.99\textwidth}{!}{%
\begin{tabular}{|p{1.8cm}|p{9.5cm}|p{5cm}|}
%\begin{tabular}{|c|c|c|}
\hline
\textbf{WEM Stability factor} & \textbf{Previous Work} & \textbf{Our Work}\\
\hline
     Datasets used (no. of tokens) & 
    $\bullet$ Wikipedia (1.5B) \cite{levy2015improving} \newline 
    $\bullet$ NYT(58M) and Europarl    (61M)\cite{wendlandt2018factors} \newline
    $\bullet$ Brown, Project Gutenberg and Reuters (10k each)\cite{chugh} \newline 
    $\bullet$ US Federal Courts of Appeals (38k), NYT (22k) and Reddit (26k)\cite{antoniak2018evaluating} \newline 
    $\bullet$ NIPS between 2007 and 2012 (2M)\cite{dridi2018k} \newline 
    $\bullet$ Ohsumed dataset (34M)\cite{lee2019understanding} \newline
    $\bullet$ Google N-gram cor-pus: English Fiction(4.8B) and German(0.7B)\cite{hellrich2016bad} \newline
    $\bullet$ BNC and ACL An-thology  Reference corpus\cite{pierrejean2018towards}
    & Wikipedia, News-Crawl (2007), Lyrics and Europarl (50M each)

    \\ \hline

    Methods studied & 
    Word2Vec \cite{wendlandt2018factors, chugh, dridi2018k, lee2019understanding, pierrejean2018towards}, GloVe \cite{levy2015improving, wendlandt2018factors, antoniak2018evaluating}, PPMI \cite{levy2015improving, wendlandt2018factors, antoniak2018evaluating}, SGNS \cite{levy2015improving, antoniak2018evaluating, hellrich2016bad}, SVD \cite{levy2015improving}, LSA \cite{antoniak2018evaluating}, SGHS \cite{hellrich2016bad}
    
    & Word2Vec, GloVe and fastText
    \\ \hline
   
    \raggedright*Number of nearest neighbors 
    & $\bullet$ Reliability while considering different nearest neighbors is very similar for all languages and time spans and WEMs considered \cite{hellrich2016bad}\newline
    $\bullet$ A few nearest neighbors are enough for stability evaluation \cite{wendlandt2018factors}
    
    & $\bullet$Stability of fastText reduces slightly with increase in the number of nearest neighbors. 
    \\ \hline
    
    Word frequency & 
    $\bullet$ Medium frequency words have the lowest reliability \cite{hellrich2016bad} \newline
    $\bullet$ Frequency is not a major factor in stability \cite{wendlandt2018factors} \newline
    $\bullet$ Words having a low and high frequency range have a tendency to display more variation. Medium frequency words show more stability \cite{pierrejean2018towards} \newline
    $\bullet$ Frequency does not directly affect the stability of medical word embeddings
    \cite{lee2019understanding}
    & $\bullet$All word groups(low, medium, and high frequency) show high variance in word stability.\newline
    $\bullet$ A single WEM need not produce the most stable embedding for each word in the vocabulary
    
    \\ \hline
    
    Number of vector dimension
    & $\bullet$ Embedding size is a causal factor and varies across corpora \cite{chugh} \newline
    $\bullet$ Stability increases considerably as the dimensionality increases, but it diminishes or becomes slightly invariant after 200 dimensions \cite{dridi2018k} 
    & $\bullet$Stability improves with increase in the vector dimensions. But the improvement plateaus after 300 dimensions
    \\ \hline
    
    Number of training epochs 
    & $\bullet$ Reliability increases for each subsequent epoch under negative sampling. SGHS has higher reliability than SGNS when trained on 1 epoch while there's no difference in terms of accuracy \cite{hellrich2016bad} \newline
    
    & $\bullet$ Stability of GloVe reduces with the number of training epochs. Effect on fastText is negligible. Stability of Word2Vec improves significantly with the number of training epochs. \\ \hline
    
    Context window size 
    & $\bullet$Larger context window size leads to better accuracy of word representations.but, it shows a diminishing return after a certain point \cite{dridi2018k} \newline
    $\bullet$ Increase in context window size gives varied results for WEMs \cite{pierrejean2018towards} \newline
    
    & $\bullet$Stability of fastText reduces slightly with increase in the context window size. Stability of Word2Vec reduces with increase in the context window size. GloVe stability increases slightly with increase in the context window size.
    $\bullet$Higher values of context window size have negligible effect on WEM stability.
    \\ \hline
    
    Downstream Tasks & Word similarity \cite{levy2015improving, wendlandt2018factors}, Word analogy \cite{levy2015improving, dridi2018k}, POS tagging \cite{wendlandt2018factors}& Word Clustering, POS Tagging and Fairness Evaluation\\ \hline
    
\end{tabular}

\caption{\label{comparisonSummary}
Summary of our findings and comparison with existing WEM stability evaluation work
}

\end{table*}

We explore the stability evaluation of three popular WEMs: Word2Vec, GloVe, and fastText. For a word $w$, we compute the stability of representing $w$ across $E_1$ and $E_2$ in terms of number of shared nearest neighbors that $E_1(w)$ and $E_2(w)$ have. Newman-Griffis and Fosler-Lussier \cite{newman2017second} have shown that nearest neighbor information alone is sufficient to capture most of the performance benefits of word embeddings. Our findings and comparison with existing stability evaluation studies is summarized in the Table \ref{comparisonSummary}. This work makes three specific contributions. First, we evaluate stability of WEMs on diverse real-world datasets. Second, we study the effect of multiple parameters on WEMs stability. Third, we analyze the effect of WEMs instability on three downstream applications.

Rest of the paper is organized as follows. Our experiment set up is described in Section 2. Our results on stability evaluation are discussed in Section 3. Effect on downstream tasks are analyzed in Section 4. Related work is covered in Section 5. Finally we conclude and discuss future work in Section 6.

%We perform experiments on four diverse real-world datasets: Wikpedia, News, Song lyrics, and European parliamentary proceedings. We analyze the effects of five parameters involved in tuning a WEM and evaluating its stability. We have selected three downstream applications: word clustering, POS tagging, and fairness measurement of word embeddings. 

\section{Experiment Set Up}
This section describes the datasets used, training methodology, and stability measurement. All code and datasets for our work are publicly available on the Web for reproducibility\footnote{https://github.com/AnganaB/Stability-of-WEMs}.

\subsection{Datasets}

We performed experiments using four publicly available real-world datasets. Please refer to Table \ref{table:Datasets Statistics} for the statistics about the datasets. These datasets represent diverse styles of natural language usage. Online collaborative writing style is covered in our Wikipedia (Sample) dataset\footnote{https://dumps.wikimedia.org/enwiki/}. It contains a subset of English Wikipedia articles from October 2017 Wikipedia dump. News articles are covered in the NewsCrawl(2007) dataset\footnote{http://www.statmt.org/wmt16/translation-task.html}. We have used the News Crawl articles for the year 2007. The Lyrics dataset represents creative writing style\footnote{Datasets from \cite{manash} and\cite{Fell2014LyricsbasedAA}}. It contains over 400K English songs of around 7.2K artists from the period 1938 to 2016. Formal parliamentary communication is covered in our Europarl (Sample) dataset\footnote{http://www.statmt.org/wmt16/translation-task.html}. It includes the English version of the proceedings of Bulgarian, Czech, Slovak, Slovene, Romanian and Polish parliaments. Our Lyrics dataset had around fifty million tokens. We sampled other datasets to match the size of our Lyrics dataset. We have deliberately chosen these diverse corpora that vary in corpus parameters like size of vocabulary and topics to assess our models with realistic examples. We have pre-processed these corpora by tokenizing sentences and removing non-alphabetic characters and converting all characters to lower cases. We have also removed the stop words by using the stop words corpora of NLTK \footnote{http://www.nltk.org/}. We have also considered a minimum word occurrence of five and removed all words that appear less than five times in a corpus.

\begin{table}[hbt]
\centering
\caption{Datasets Statistics.}
\label{table:Datasets Statistics}
\begin{tabular}{|c|c|c|}
\hline
  \bf Dataset & \bf Vocab Size &\bf No. of Tokens     \\
\hline

Wikipedia (Sample) & 216,580 & 46,359,850       \\
\hline
News Crawl (2007) & 116,401 & 45,149,886      \\
\hline
Lyrics & 85,378 & 49,127,358  
   \\
\hline
Europarl (Sample) & 41,674 & 41,330,440         \\
\hline

\end{tabular}
\end{table}

%Please refer to Table~\ref{table:Datasets Statistics} for the statistics of our datasets. For our work, we gathered data from four sources namely   Wikipedia \footnote{https://dumps.wikimedia.org/enwiki/}, News Crawl Articles \footnote{http://www.statmt.org/wmt16/translation-task.html}, Lyrics \cite{manash,Fell2014LyricsbasedAA} and Europarl \cite{Koehn_europarl:a}. The Wikipedia (sample) corpus contains a subset of the Wikipedia articles till October 2017.  The lyrics corpus contains over 400K English songs of around 7.2K artists from 1938 to 2016. Our Europarl corpus sample includes the English version of the proceedings of Bulgarian, Czech, Slovak, Slovene, Romanian and Polish parliaments. We have deliberately chosen these diverse corpora that vary in corpus parameters like size of vocabulary and topics to assess our models with realistic examples. We have pre-processed these corpora by tokenizing sentences and removing non-alphabetic characters and converting all characters to lower cases. We have also removed the stop words by using the stop words corpora of NLTK \footnote{http://www.nltk.org/}. We have also considered a minimum word occurrence of 5 and removed all words that appear less than five times in a corpus. 

\subsection{Training}
Evaluating all the WEMs that are currently available is beyond the scope of this work. We focused on three popular WEMs: Word2Vec, GloVe, and fastText. For Word2Vec, we used the Continuous Bag of Words (CBOW) variant from the Gensim \footnote{https://github.com/RaRe-Technologies/gensim/tree/develop/gensim/} Python library to train our models. For the fastText, we used both the original fastText \footnote{https://github.com/facebookresearch/fastText}  library and its Gensim version. The original fastText library does not provide the option for setting a random seed value, which is provided by the Gensim version. We trained the GloVe models using the original GloVe code \footnote{https://github.com/stanfordnlp/GloVe} with a slight modification to input random seed as a parameter. For all the models trained on all the three WEMs, we experimented with the vector dimensions, iterations, and window size, keeping all other parameters at default settings. For all the experiments on a corpus, we measured stability across five randomized embedding spaces trained with an algorithm and took the average among them. We used the same random seeds for all the WEMs.

\subsection{Stability Evaluation}
\label{stabilityEval}
Consider a dataset $C$ with vocabulary of size $VocabSize$. A WEM $A$ is used to learn word representations in an embeddings space using the dataset $C$. Let $E_1$ and $E_2$ be two such embedding spaces. These two embedding spaces correspond to two distinct executions of WEM $A$ over the dataset $C$. These two executions of $A$ can differ from each other in terms of one or multiple parameters. We evaluate the stability of $A$ across $E_1$ and $E_2$ as follows
\begin{align}
Stability (A) = \frac{\sum_{i=1}^{VocabSize}stability (w_{i})}{VocabSize} 
\end{align}

Stability of each word $w$ in the vocabulary is computed as follows
\begin{align}
stability (w) = \frac{\left|KNN_{E_1} (w)\bigcap KNN_{E_2} (w)\right|}{k} 
\end{align}

$KNN_{E}(w)$ represents the set of top $k$ nearest neighbors of word $w$ in the embedding space $E$. These nearest neighbors are computed using the cosine similarity between the word embeddings. Existing stability evaluation studies use the similar method for stability computation.

\section{Results}
This section describes our experimental results on stability evaluation. We have experimented with five parameters that can affect the stability evaluation. For each such parameter,we will present results here only for one dataset due to the space limitation. However, we have observed similar trends across all four datasets. These additional experimental results are available along with our code.

\subsection{Number of Nearest Neighbors}

\begin{figure}
     \centering
     \begin{subfigure}[b]{0.4\textwidth}
         \centering
         \includegraphics[width=\textwidth]{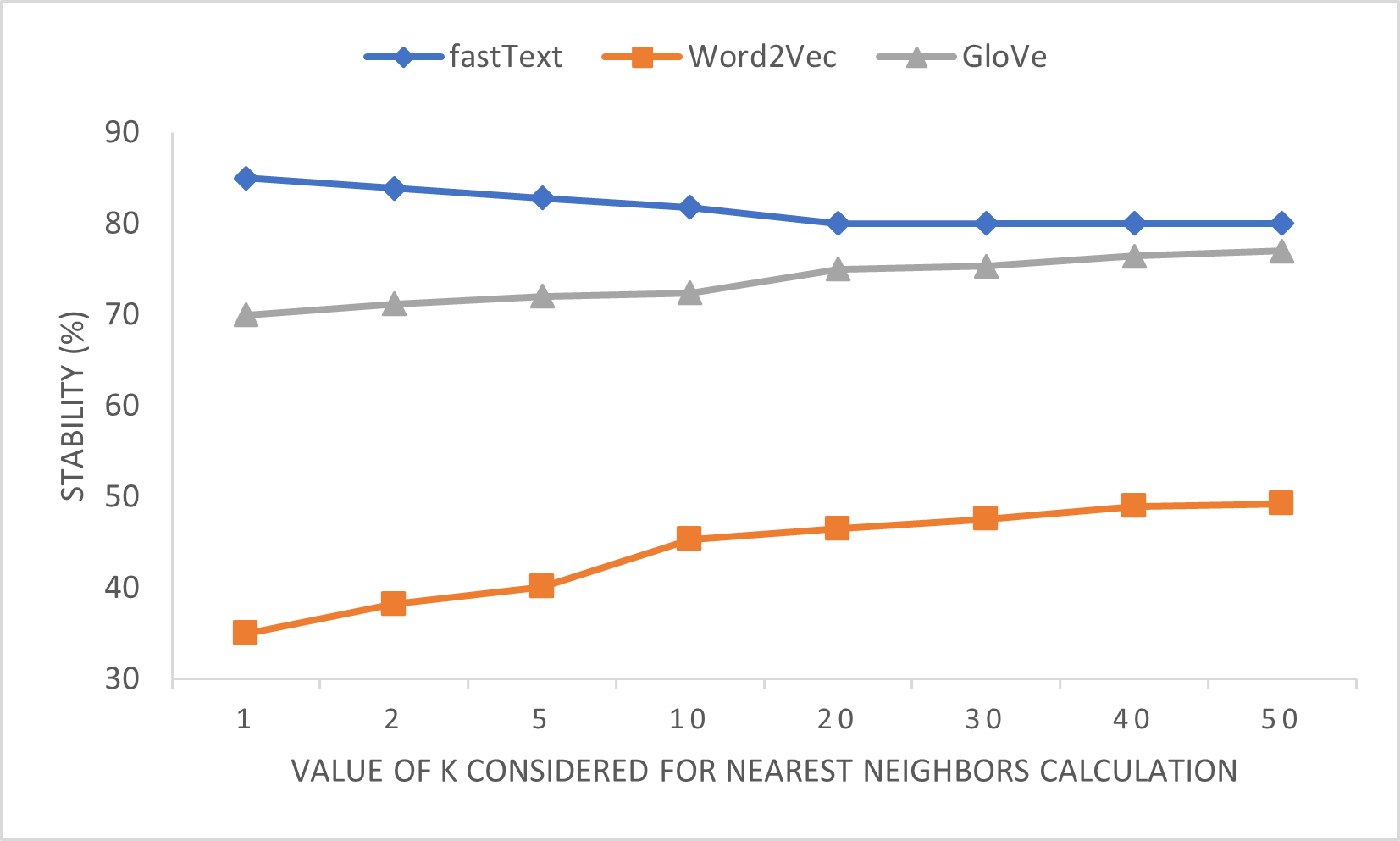}
         \caption{Wikipedia}
         \label{fig:6}
     \end{subfigure}
     \hfill
     \begin{subfigure}[b]{0.4\textwidth}
         \centering
         \includegraphics[width=\textwidth]{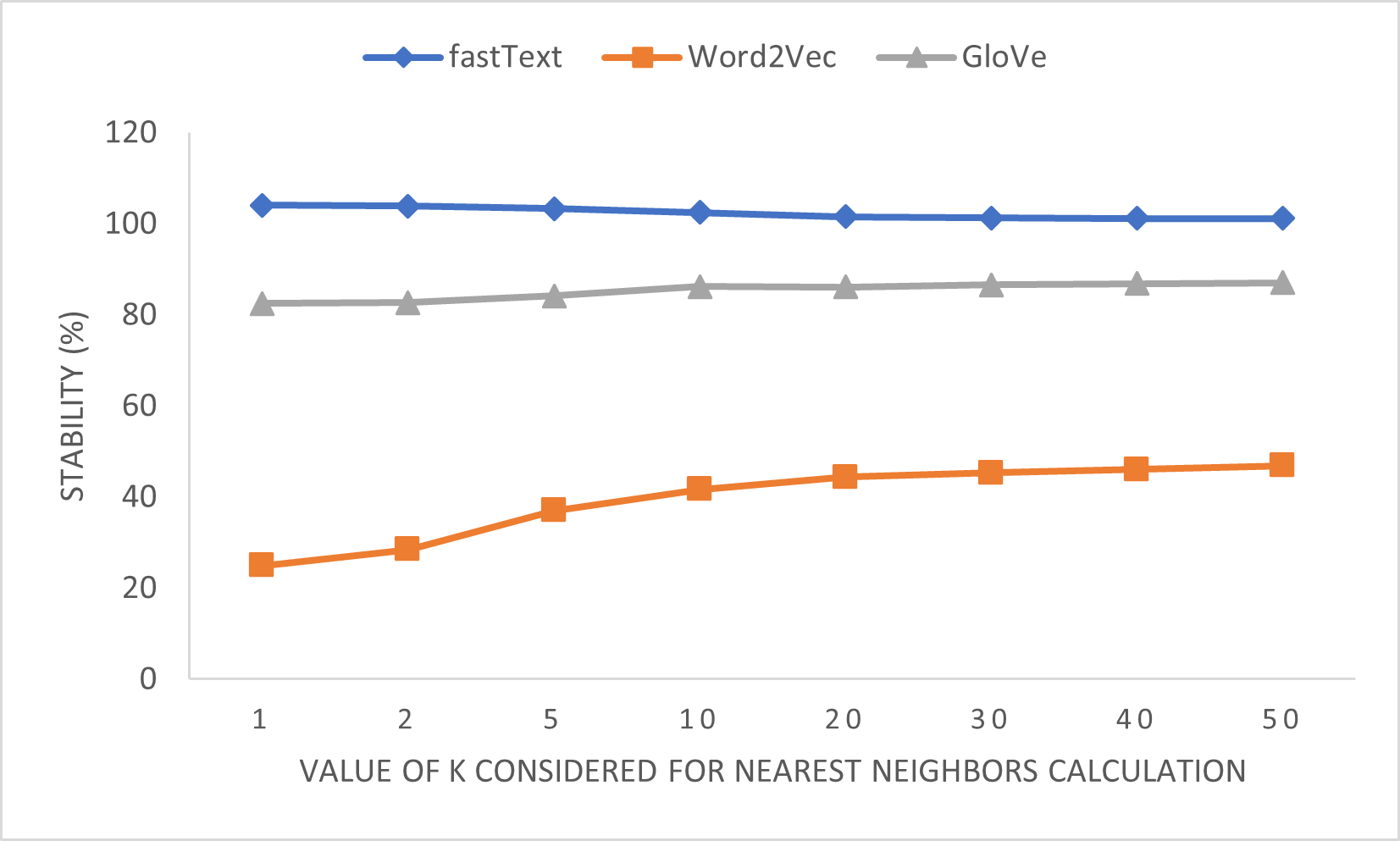}
         \caption{Europarl}
         \label{fig:ae}
     \end{subfigure}
     \hfill
     \begin{subfigure}[b]{0.4\textwidth}
         \centering
         \includegraphics[width=\textwidth]{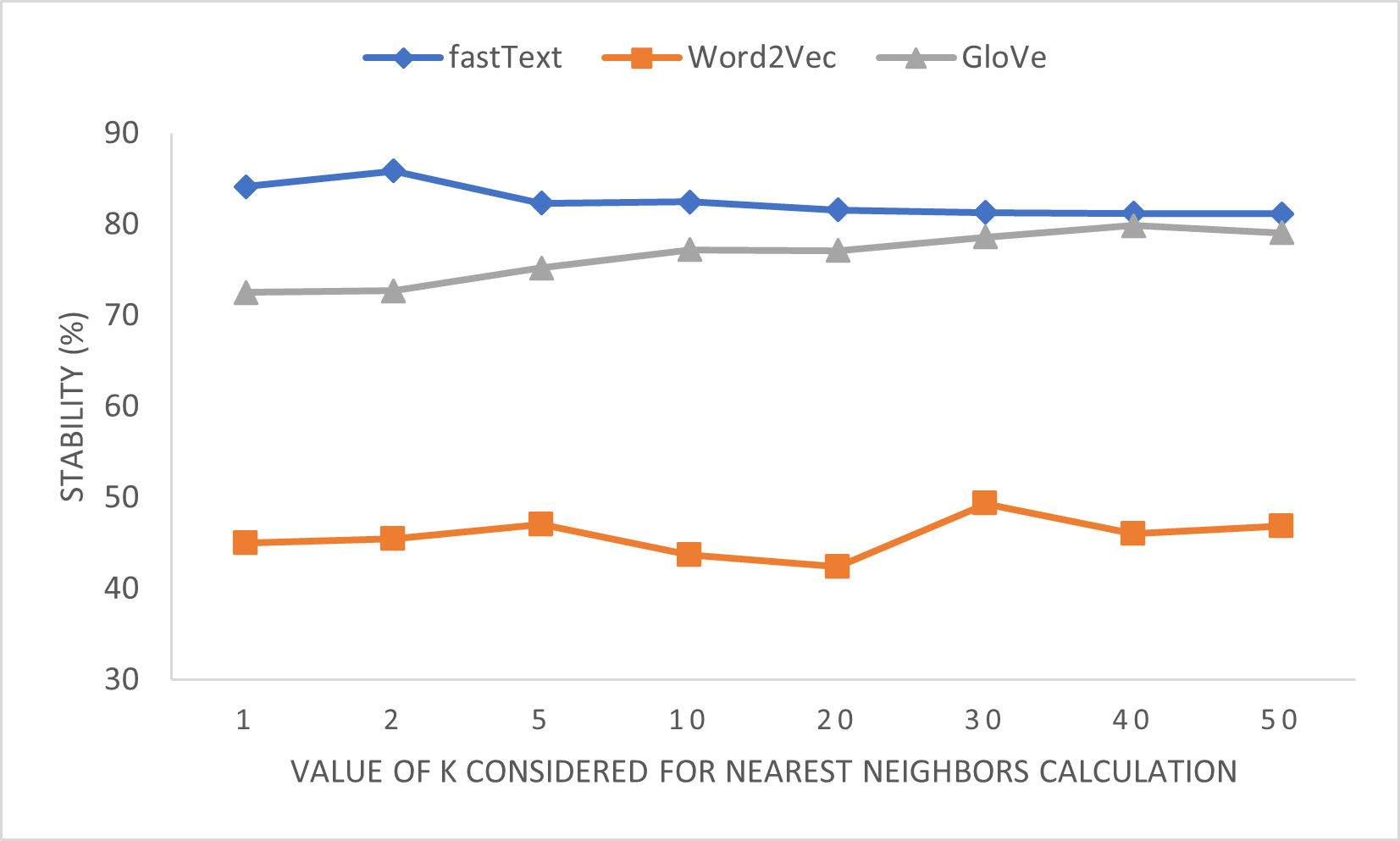}
         \caption{Lyrics}
         \label{fig:b}
     \end{subfigure}
     \hfill
     \begin{subfigure}[b]{0.4\textwidth}
         \centering
         \includegraphics[width=\textwidth]{nn_wiki.png}
         \caption{News 2007}
         \label{fig:c}
     \end{subfigure}
        \caption{WEM stability plateaus for higher values of $k$}
        \label{nnEffect}
\end{figure}

The word stability computation formula in the Section \ref{stabilityEval} involves a parameter $k$. It indicates the number of nearest neighbors considered for stability evaluation. If the value of $k$ is set close to one then it is too restrictive and we expect the stability value for WEMs to be low. As we increase the value of $k$, we expect the stability to increase. However, with very high value of $k$, the stability evaluation will be meaningless. For example, in the extreme case of $k=VocabSize-1$, stability of any WEM will be hundred percent.

Please refer to Figure \ref{nnEffect}. This figure shows the variation in the stability of WEMs with respect to the value of $k$.We can observe that for Word2Vec and GloVe, the stability increases with increase in the value of $k$. However, it plateaus after $k=10$. This observation correlates with existing sability evaluation studies\cite{wendlandt2018factors}. In contrast, we observe unexpected behavior for the fastText. With increase in the value of $k$, its stability actually reduces slightly. This indicates that fastText maintains the top nearest neighbors across the embedding spaces. However, fastText fails to maintain the stability for the lower ranked nearest neighbors. Still, fastText is the most stable WEM amongst the three.

\subsection{Word Frequency}
Consider a dataset $C$ with the vocabulary of size $VocabSize$. To observe the effect of word frequency on word stability, we divide the vocabulary into five groups of equal size ($VocabSize/5$). Based on the word frequency, these groups are named as : Very Low Frequency ($VL$), Low Frequency ($L$), Medium Frequency ($M$), High Frequency ($H$), and Very High Frequency ($VH$). Each group accounts for twenty percent of the words ranked by frequency. For example, the $VH$ group will have the top twenty percent most frequent words from $C$ and the $VL$ group will have the bottom twenty percent frequent words from $C$.

\begin{figure}
     \centering
     \begin{subfigure}[b]{0.25\textwidth}
         \centering
         \includegraphics[width=\textwidth]{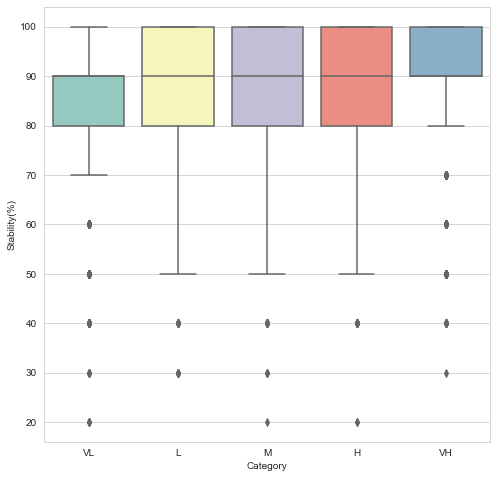}
         \caption{Wikipedia}
         \label{fig:1}
     \end{subfigure}
     \hfill
     \begin{subfigure}[b]{0.25\textwidth}
         \centering
         \includegraphics[width=\textwidth]{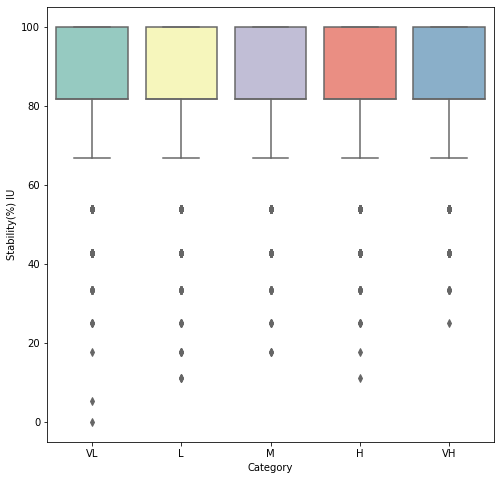}
         \caption{Europarl}
         \label{fig:af}
     \end{subfigure}
     \hfill
     \begin{subfigure}[b]{0.25\textwidth}
         \centering
         \includegraphics[width=\textwidth]{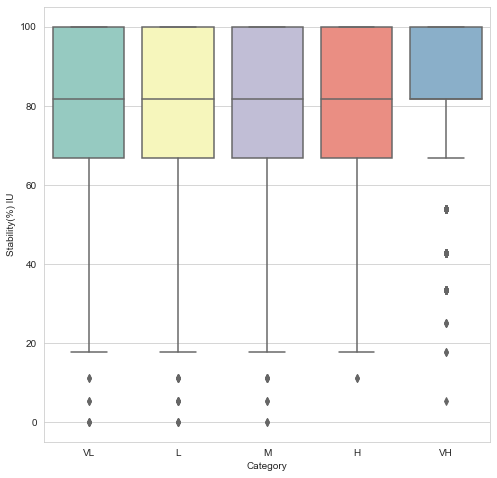}
         \caption{Lyrics}
         \label{fig:d}
     \end{subfigure}
     \hfill
     \begin{subfigure}[b]{0.25\textwidth}
         \centering
         \includegraphics[width=\textwidth]{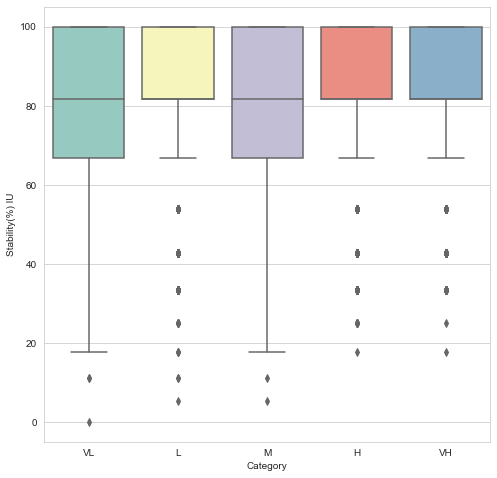}
         \caption{News 2007}
         \label{fig:e}
     \end{subfigure}

        \caption{All word groups have high variance in word stability for fastText}
        \label{freqEffect1}
\end{figure}

\begin{figure}
     \centering    
     
     \begin{subfigure}[b]{0.25\textwidth}
         \centering
         \includegraphics[width=\textwidth]{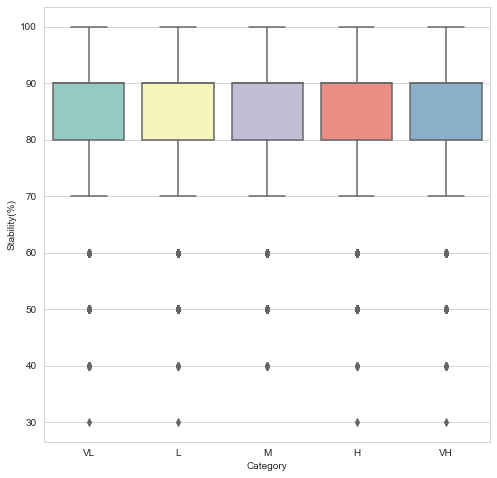}
         \caption{Wikipedia}
         \label{fig:ag}
     \end{subfigure}
     \hfill
     \begin{subfigure}[b]{0.25\textwidth}
         \centering
         \includegraphics[width=\textwidth]{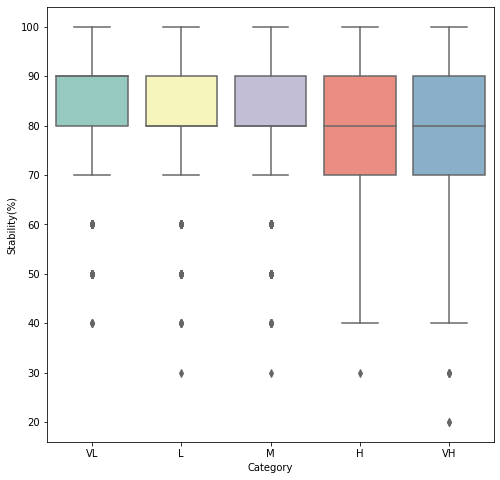}
         \caption{Europarl}
         \label{fig:f}
     \end{subfigure}
     \hfill
     \begin{subfigure}[b]{0.25\textwidth}
         \centering
         \includegraphics[width=\textwidth]{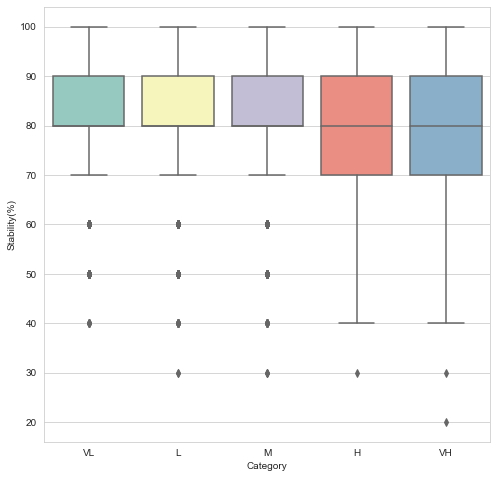}
         \caption{Lyrics}
         \label{fig:g}
     \end{subfigure}
     \hfill
     \begin{subfigure}[b]{0.25\textwidth}
         \centering
         \includegraphics[width=\textwidth]{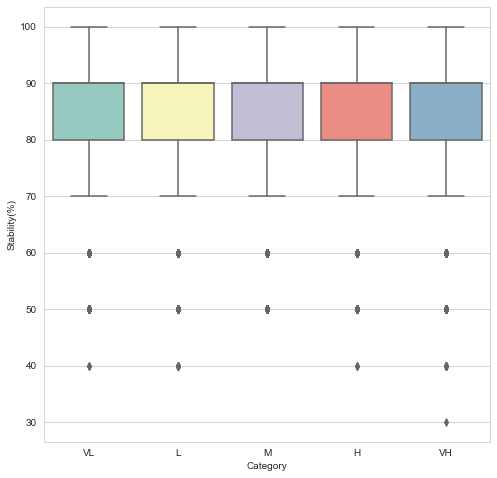}
         \caption{News 2007}
         \label{fig:ah}
     \end{subfigure}

\caption{All word groups have high variance in word stability for GloVe}
        \label{freqEffect2}
\end{figure}

\begin{figure}
     \centering    
     
     \begin{subfigure}[b]{0.25\textwidth}
         \centering
         \includegraphics[width=\textwidth]{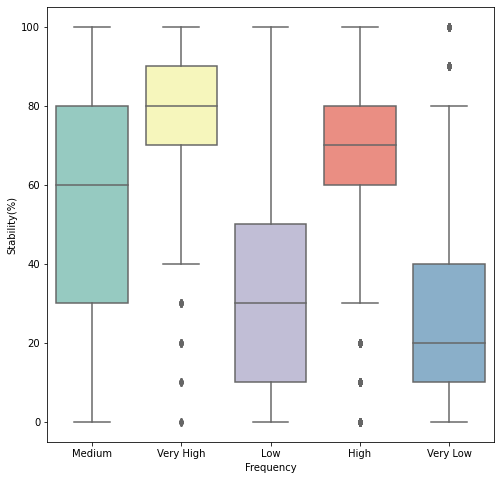}
         \caption{Wikipedia}
         \label{fig:h}
     \end{subfigure}
     \hfill
     \begin{subfigure}[b]{0.25\textwidth}
         \centering
         \includegraphics[width=\textwidth]{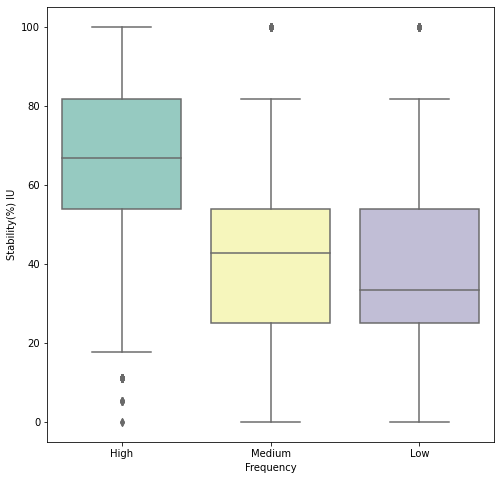}
         \caption{Europarl}
         \label{fig:i}
     \end{subfigure}
     \hfill
     \begin{subfigure}[b]{0.25\textwidth}
         \centering
         \includegraphics[width=\textwidth]{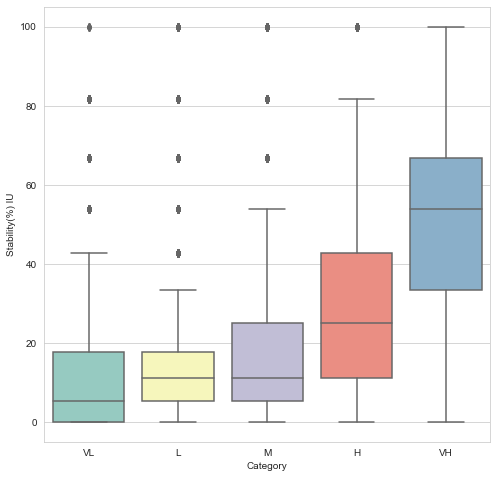}
         \caption{Lyrics}
         \label{fig:j}
     \end{subfigure}
     \hfill
     \begin{subfigure}[b]{0.25\textwidth}
         \centering
         \includegraphics[width=\textwidth]{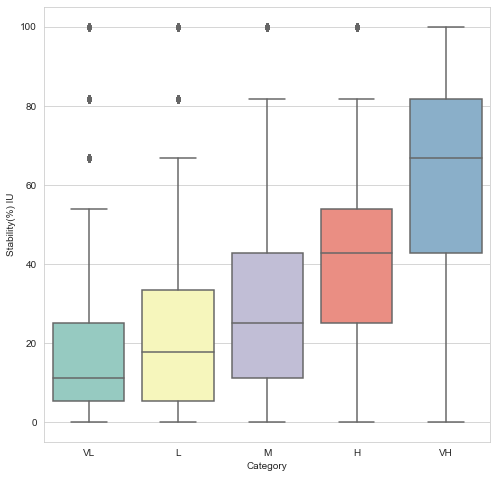}
         \caption{News 2007}
         \label{fig:k}
     \end{subfigure}
        \caption{All word groups have high variance in word stability for Word2Vec.}
        \label{freqEffect3}
\end{figure}

Please refer to Figures \ref{freqEffect1}, \ref{freqEffect2} \ref{freqEffect3}. These figures show the boxplot for word stability when words are grouped based on frequency as described before. Previous studies have reported that stability of a word group increases with increase in the average frequency of the group \cite{wendlandt2018factors}. Which means that as we move from $VL$ group to $VH$ group, the average or median word stability should increase. Our experimental results correlate with the existing studies. However, Wendlandt et al also report low variance in word stability for low as well as high frequency word groups \cite{wendlandt2018factors}. They report high variance in word stability only for medium frequency word groups. In contrast, we observe that all word groups have high variance in word stability.

\begin{figure}
     \centering
     \begin{subfigure}[b]{0.4\textwidth}
         \centering
         \includegraphics[width=\textwidth]{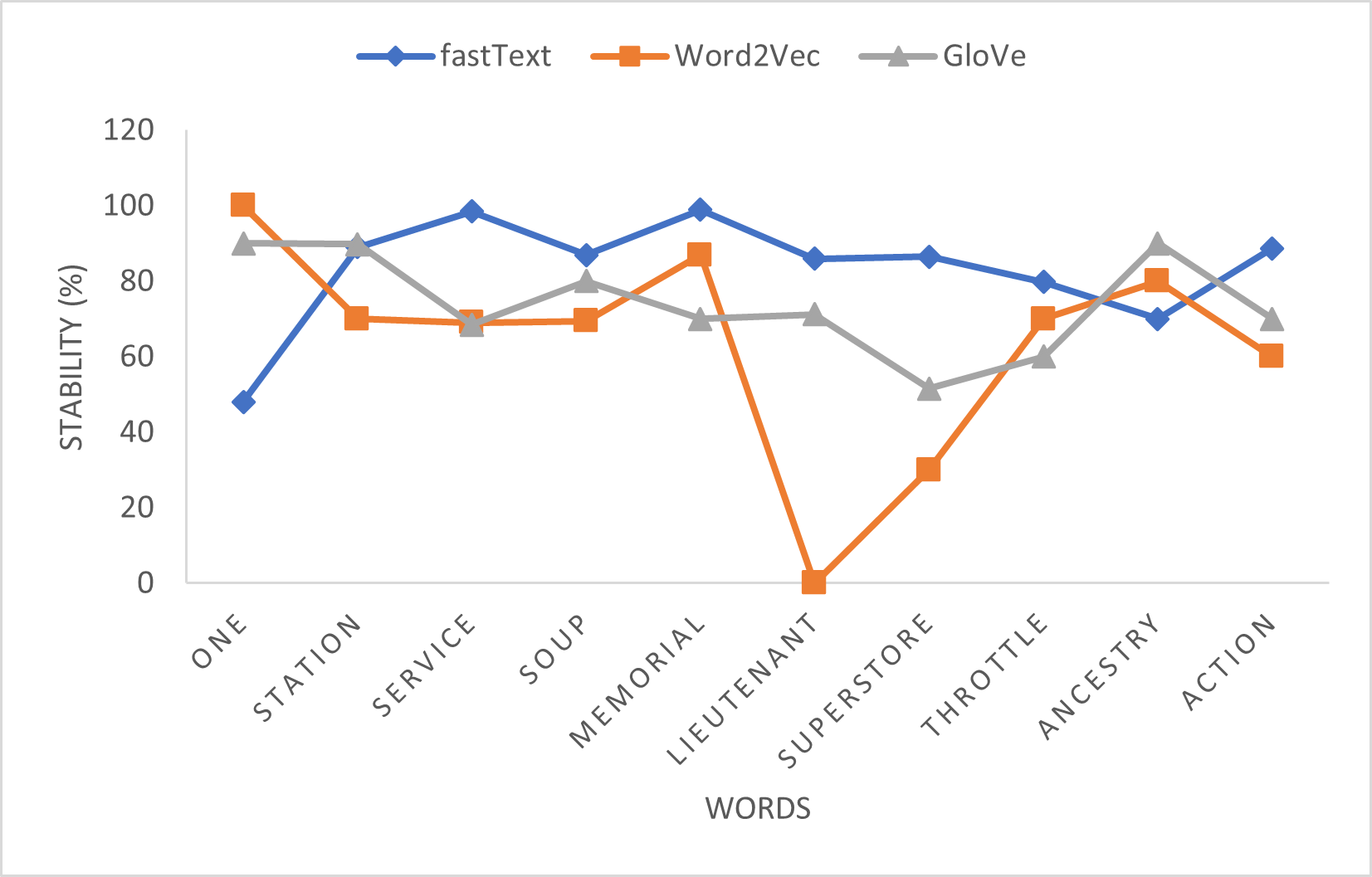}
         \caption{Wikipedia}
         \label{wordwiki}
     \end{subfigure}
     \hfill
     \begin{subfigure}[b]{0.4\textwidth}
         \centering
         \includegraphics[width=\textwidth]{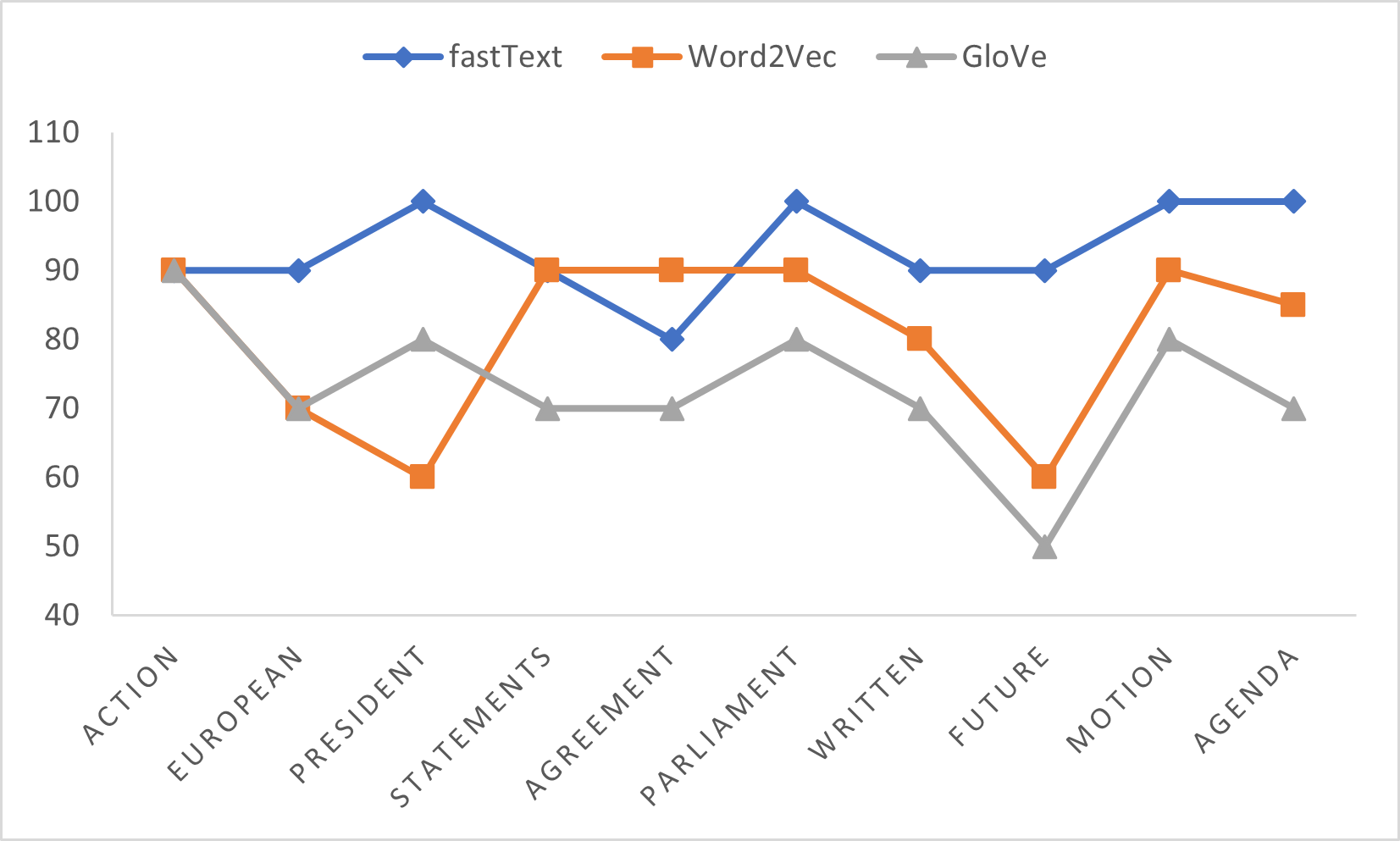}
         \caption{Europarl}
         \label{fig:ai}
     \end{subfigure}
     \hfill
     \begin{subfigure}[b]{0.4\textwidth}
         \centering
         \includegraphics[width=\textwidth]{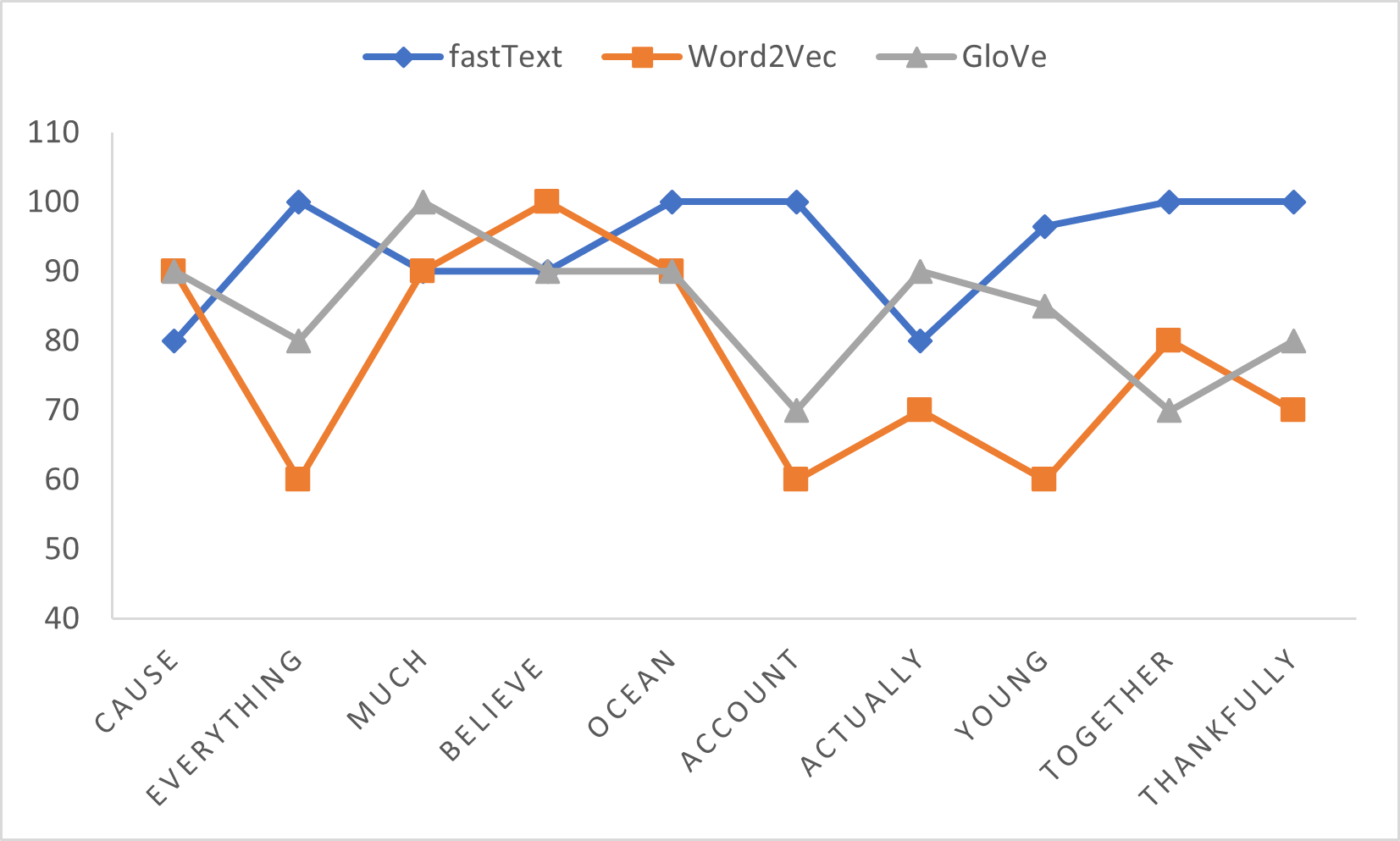}
         \caption{Lyrics}
         \label{fig:l}
     \end{subfigure}
     \hfill
     \begin{subfigure}[b]{0.4\textwidth}
         \centering
         \includegraphics[width=\textwidth]{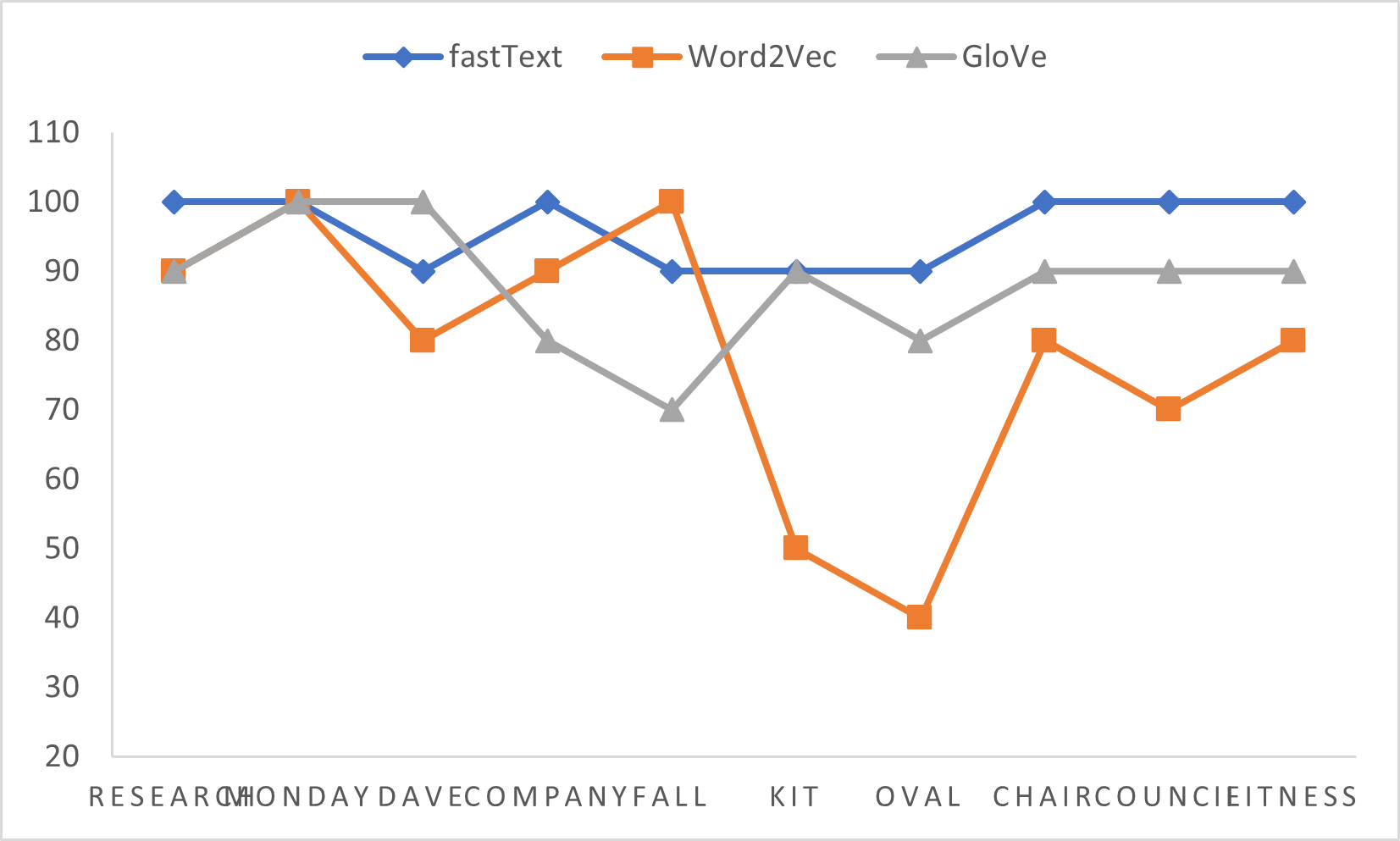}
         \caption{News 2007}
         \label{fig:m}
     \end{subfigure}
        \caption{For a set of randomly sampled words, stability of words with each WEM}
        \label{word}
\end{figure}

We can also observe from Figure~\ref{word} that the stability of a word varies with word embedding methods used. The words in Figure~\ref{wordwiki} were randomly picked from the Wikipedia (sample) corpus and these words have different frequencies ranging from 7 occurrences for "lieutenant" to 174163 for "one". To generalize this observation to the whole vocabulary, we computed the stability of each word using three WEMs. For each word $w$, now we had three stability values. For the word $w$, we chose the WEM with highest stability value amongst the three values. We observed that out of all words, each WEM was chosen for the following percentage of words: fastText (63\%), GloVe (24\%), and Word2Vec (13\%).  This implies that fastText is in general a more stable WEM. However, a particular WEM need not be the most stable method for each word in the vocabulary.

\begin{comment}

\begin{table}[hbt]
\centering
\caption{Cumulative Average Stability.}
\label{table:cum_stab}
\begin{tabular}{|c|c|c|c|}
\hline
  \bf Dataset & \bf GloVe &\bf W2V   &\bf FT  \\
\hline

Wikipedia (Sample) & 0.63 & 0.56    & 0.71   \\
\hline
News Crawl (2007) & 0.60 & 0.59 & 0.70     \\
\hline
Lyrics & 0.45 & 0.54 & 0.71
   \\
\hline
Europarl (Sample) & 0.47 & 0.60 & 0.37         \\
\hline

\end{tabular}
\end{table}

\begin{table}[hbt]
\centering
\caption{Correlation Assessment of Stability of Words based on Frequency across Algorithms.}
\label{table:corr_stab}
\begin{tabular}{|c|c|c|c|}
\hline
  \bf Dataset & \bf FT-W2V &\bf GloVe-W2V   &\bf FT-GloVe  \\
\hline

Wikipedia (Sample) & 0.95 & 0.94    & 0.16   \\
\hline
News Crawl (2007) & 0.90 & 0.88 & 0.21     \\
\hline
Lyrics & 0.94 & 0.84 & 0.64
   \\
\hline
Europarl (Sample) & 0.87 & 0.67 & 0.49         \\
\hline

\end{tabular}
\end{table}

\end{comment}

\subsection{Number of Vector Dimensions}
\begin{figure}
     \centering
     \begin{subfigure}[b]{0.4\textwidth}
         \centering
         \includegraphics[width=\textwidth]{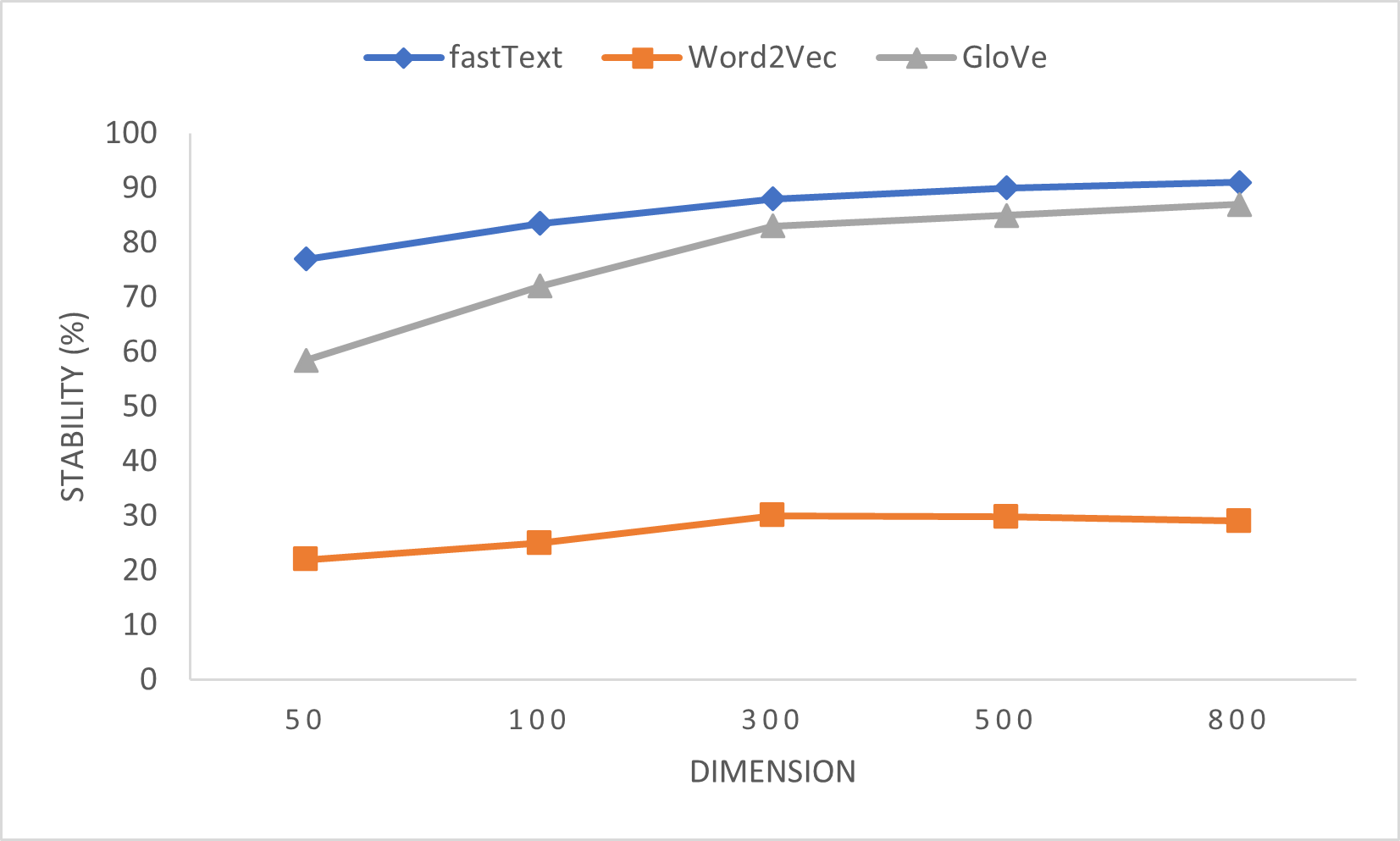}
         \caption{Wikipedia}
         \label{fig:7}
     \end{subfigure}
     \hfill
     \begin{subfigure}[b]{0.4\textwidth}
         \centering
         \includegraphics[width=\textwidth]{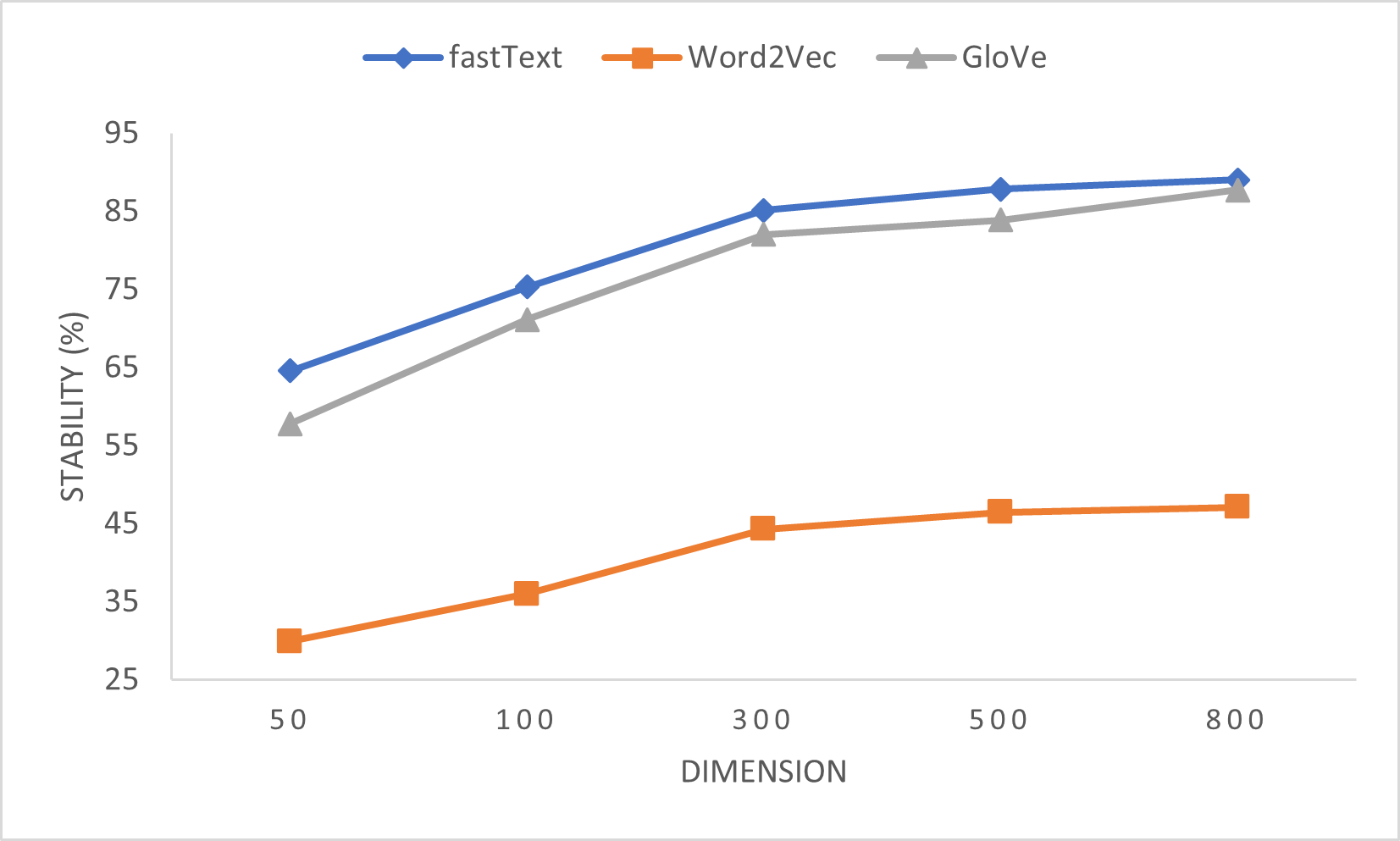}
         \caption{Europarl}
         \label{fig:aj}
     \end{subfigure}
     \hfill
     \begin{subfigure}[b]{0.4\textwidth}
         \centering
         \includegraphics[width=\textwidth]{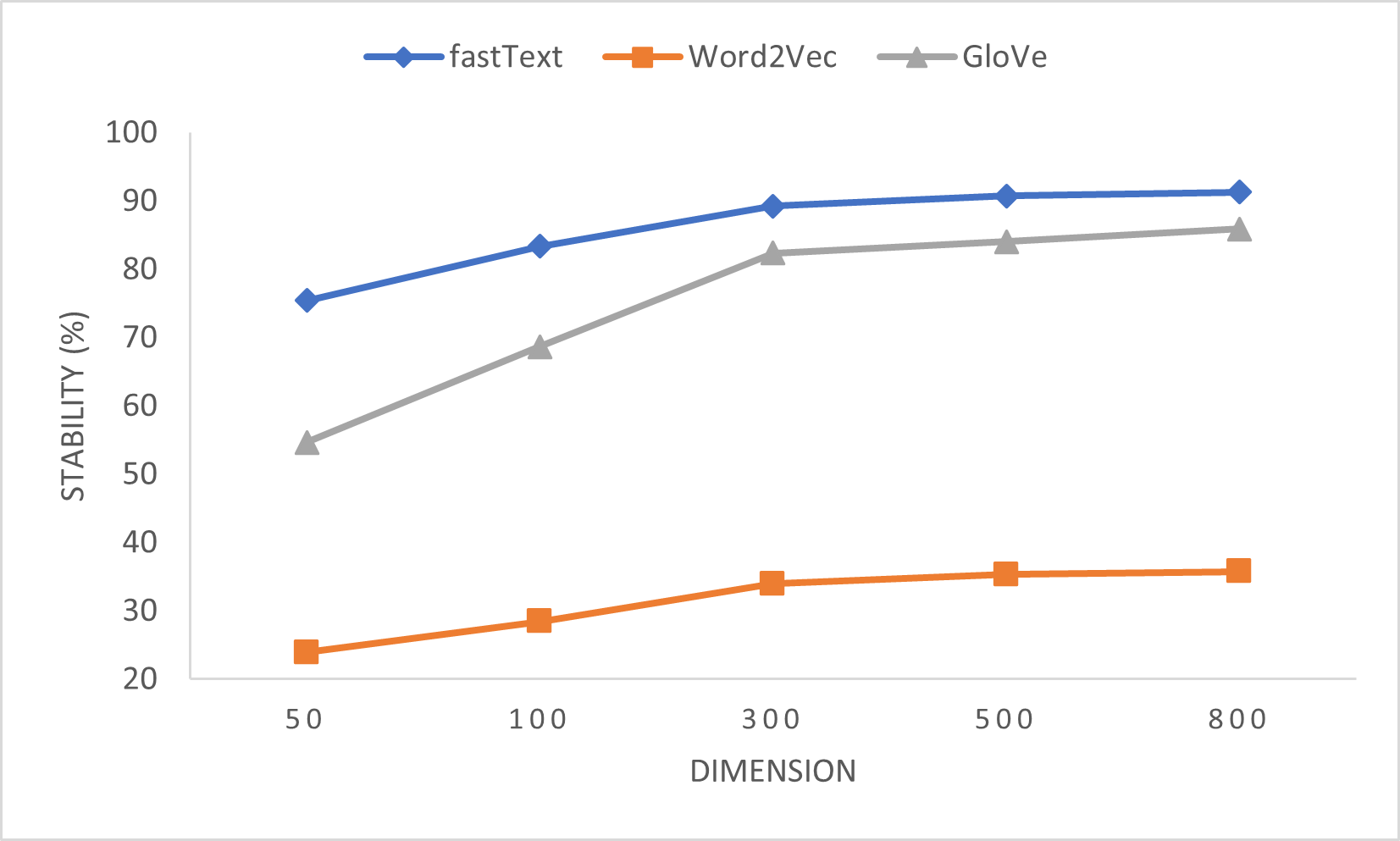}
         \caption{Lyrics}
         \label{fig:n}
     \end{subfigure}
     \hfill
     \begin{subfigure}[b]{0.4\textwidth}
         \centering
         \includegraphics[width=\textwidth]{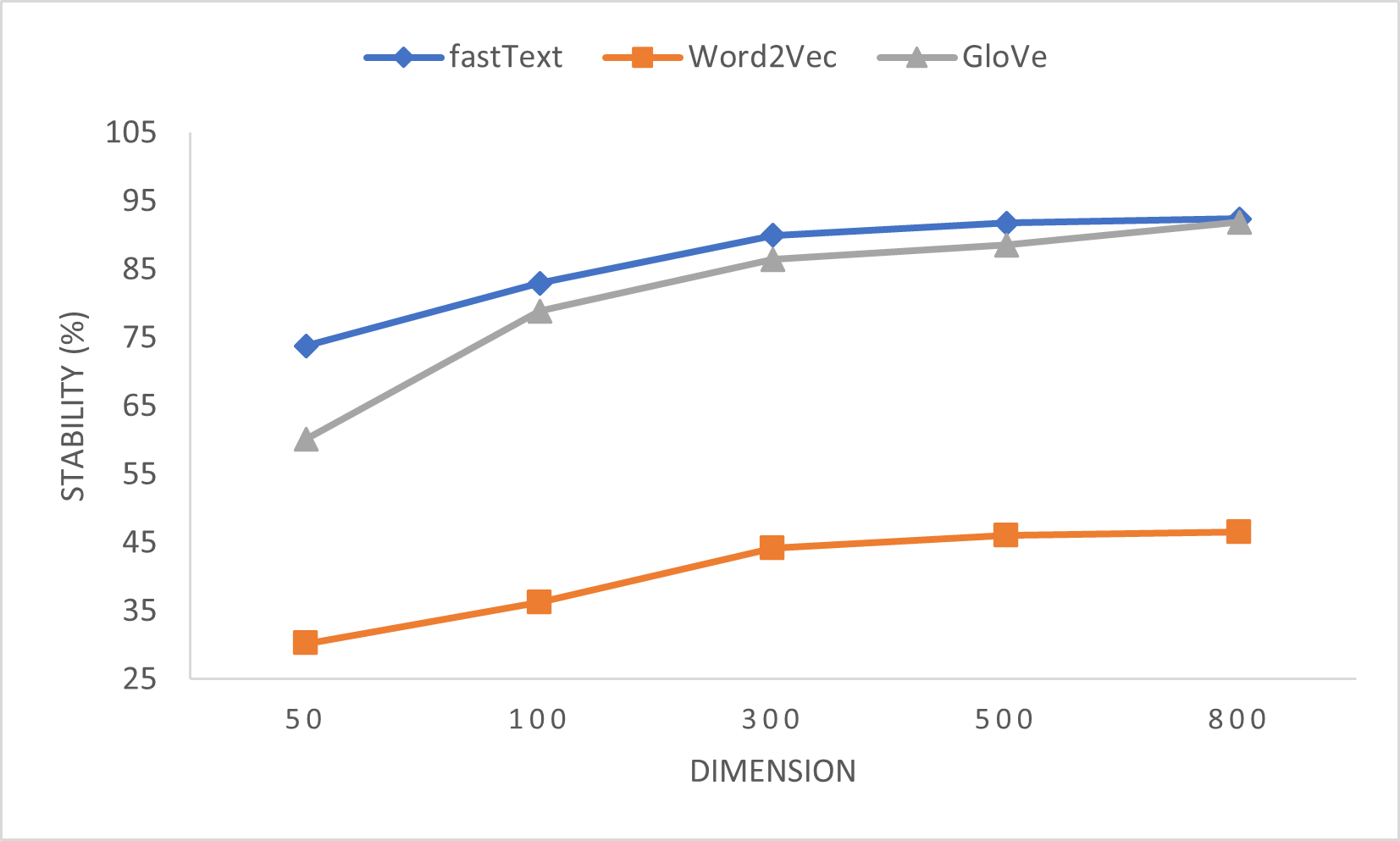}
         \caption{News 2007}
         \label{fig:o}
     \end{subfigure}
        \caption{Stability improvement plateaus after 300 dimensions}
        \label{swiki}
\end{figure}

One of the design choices with word embeddings is the number of dimensions. Lower number of dimensions reduce the computation but they might fail to represent all semantic relations. On the other hand, higher number of dimensions increase the computation while providing better semantic representation. Here we explore this design choice from the stability point of view. Figure \ref{swiki} shows variation in stability with respect to the number of dimensions for word embeddings. For each WEM, we are comparing stability between two embedding spaces with the same dimensionality. These two embedding spaces differ from each other only in terms of the initial random seeds. For example, two embedding spaces of 300 dimensioanlity  that are trained using fastText, have stability of eighty percent. As expected, the stability improves with increase in the number of dimensions. However, the improvement plateaus after 300 dimensions for all WEMs. Stability of fastText is higher than GloVe and Word2Vec for all dimension sizes. Stability of GloVe is lower for lower dimensions but increases rapidly till dimension 300. After dimension 300, the stability of GloVe becomes comparable to fastText. Stability for Word2Vec is the lowest compared to the other WEMs. We would recommend users to train their models using a vector dimension of around 300 for optimal stability.

\begin{figure}
     \centering
     \begin{subfigure}[b]{0.4\textwidth}
         \centering
         \includegraphics[width=\textwidth]{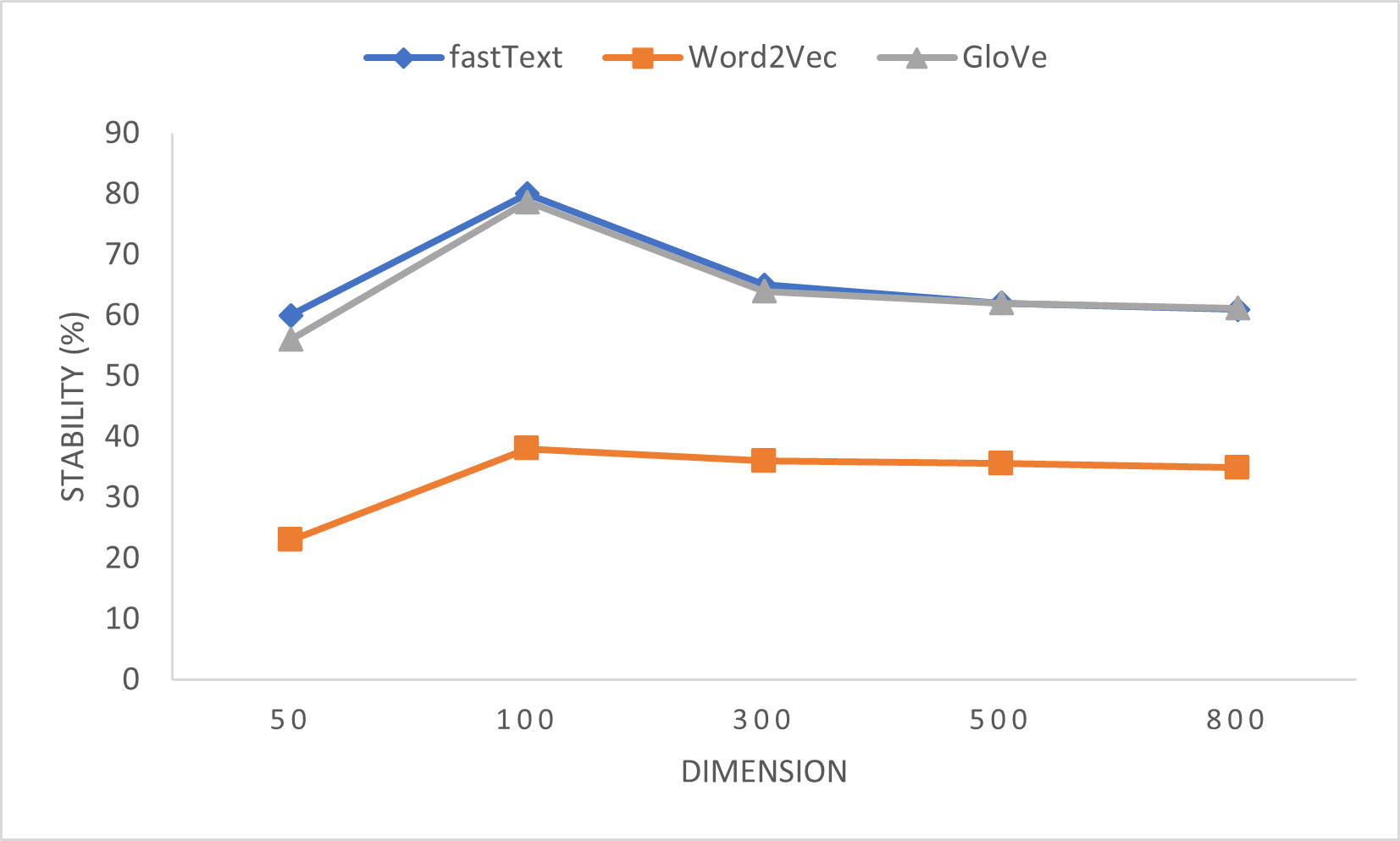}
         \caption{Wikipedia}
         \label{fig:2}
     \end{subfigure}
     \hfill
     \begin{subfigure}[b]{0.4\textwidth}
         \centering
         \includegraphics[width=\textwidth]{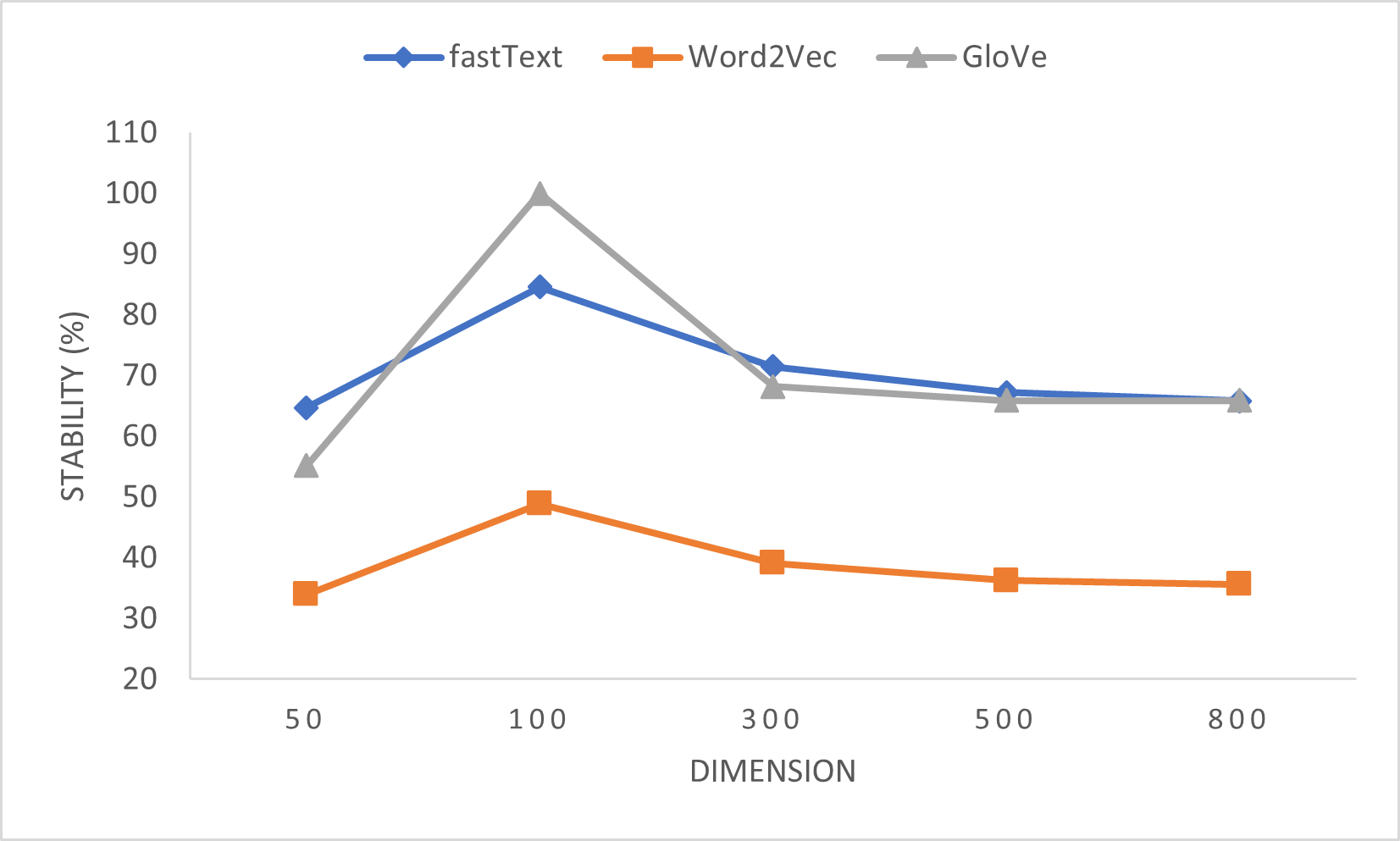}
         \caption{Europarl}
         \label{fig:aa}
     \end{subfigure}
     \hfill
     \begin{subfigure}[b]{0.4\textwidth}
         \centering
         \includegraphics[width=\textwidth]{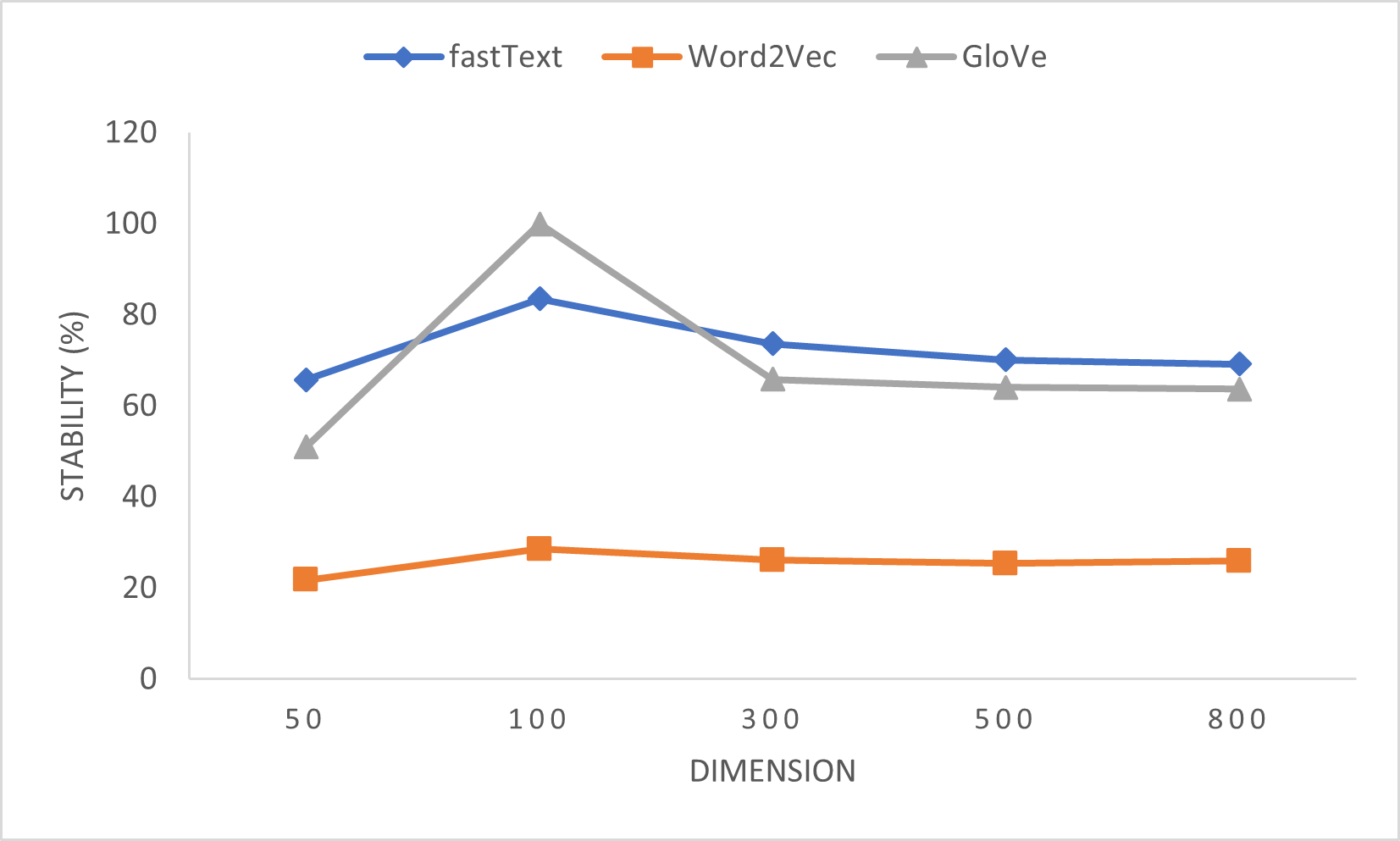}
         \caption{Lyrics}
         \label{fig:p}
     \end{subfigure}
     \hfill
     \begin{subfigure}[b]{0.4\textwidth}
         \centering
         \includegraphics[width=\textwidth]{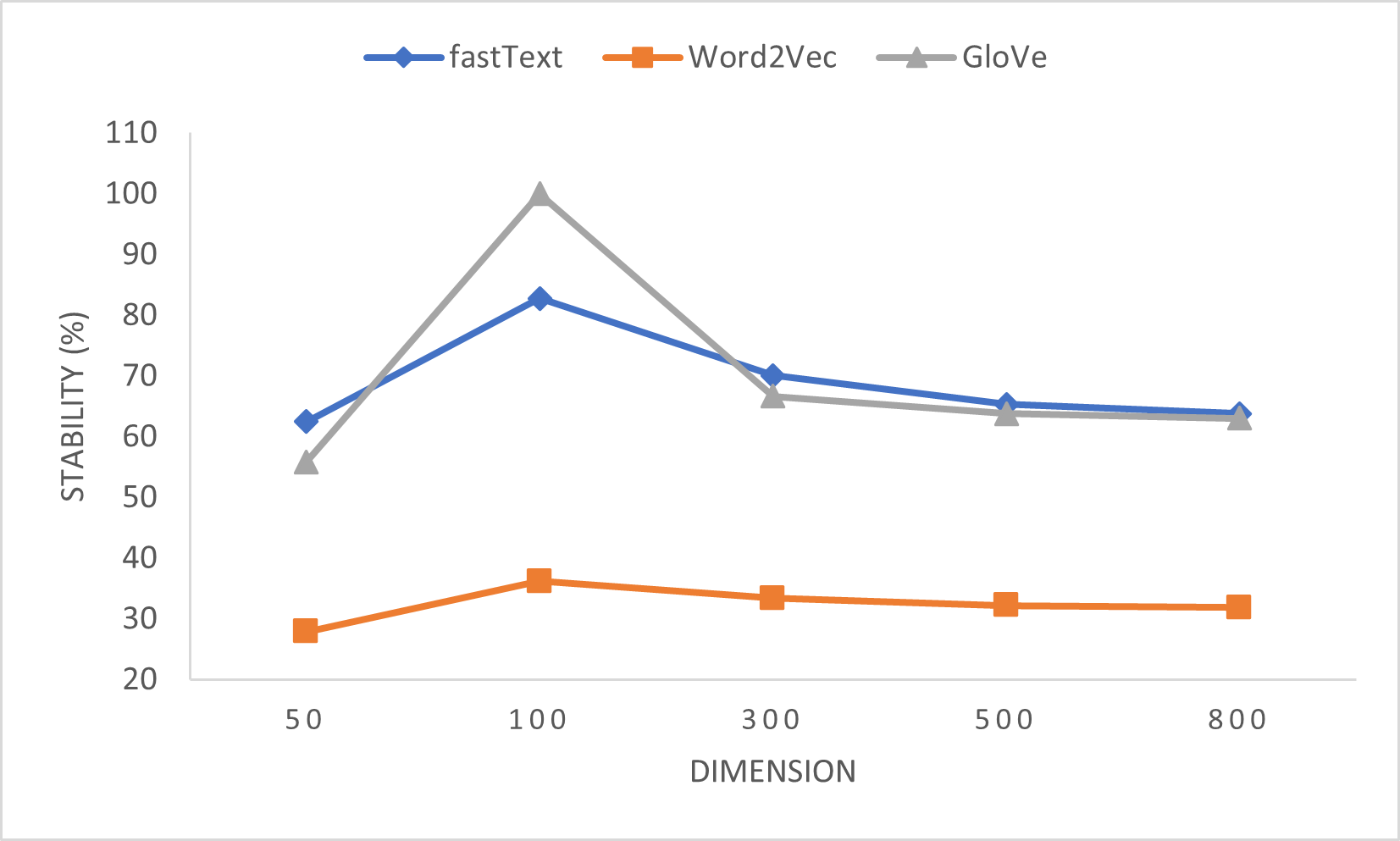}
         \caption{News 2007}
         \label{fig:q}
     \end{subfigure}
        \caption{Stability between embedding spaces $E_1$ and $E_2$. $Dim(E_1)=100$. X-axis represents $Dim(E_2)$}
        \label{across_dim}
\end{figure}

Please refer to Figure \ref{across_dim}. Here we are computing stability between embedding spaces of different dimensionality. For each WEM, consider two embedding spaces $E_1$ and $E_2$. Let $Dim(E_1)$ and $Dim(E_2)$ represent dimensionality of the respective embedding spaces. We set $Dim(E_1)$ to 100. We vary $Dim(E_2)$ from 50 to 800. We can observe that for all WEMs, stability values peaks when $Dim(E_2)$ is 100. It is expected as at $Dim(E_2)=100$, we are comparing embedding spaces of the same dimensionality. For $Dim(E_2)$, as we move away from 100, the stability decreases. This indicates that as we change the dimensionality of embedding spaces, the underlying captured semantics also change. The decrease in stability is rapid for lower number of dimensions. Probable reason for this rapid decrease is that we lose a lot of semantic information as we move to the lower number of dimensions. However, stability reduction is not that rapid as we move towards higher dimensions. We can observe only slight reduction in stability from dimension size 300 to 800. With higher dimensions, we do capture new semantic information. However, with higher dimension size, the rate of growth of semantic information also slows down. This might be a probable reason for slow reduction in stability for higher dimension size.

\subsection{Number of Training Epochs}

\begin{figure}
     \centering
     \begin{subfigure}[b]{0.4\textwidth}
         \centering
         \includegraphics[width=\textwidth]{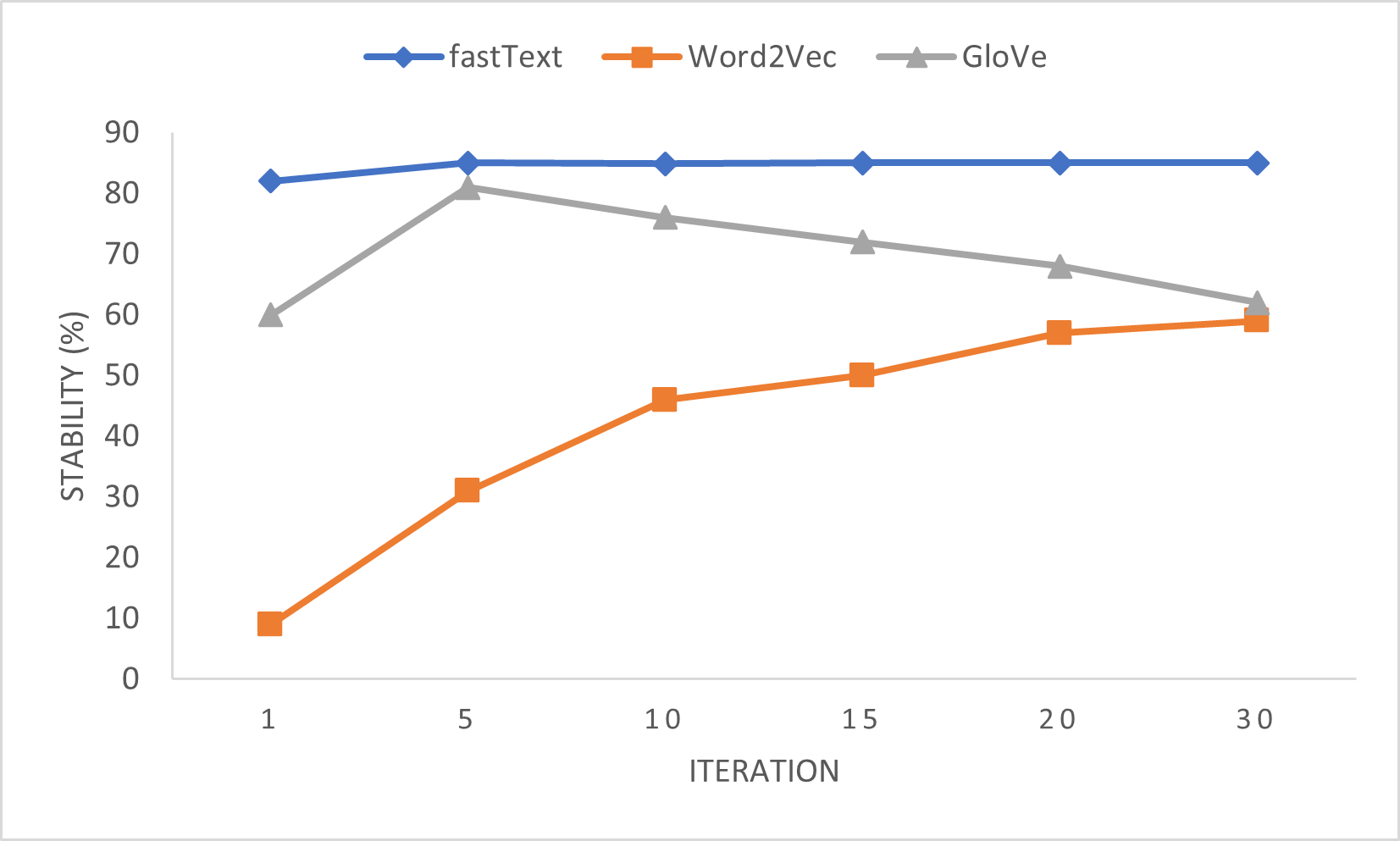}
         \caption{Wikipedia}
         \label{fig:3}
     \end{subfigure}
     \hfill
     \begin{subfigure}[b]{0.4\textwidth}
         \centering
         \includegraphics[width=\textwidth]{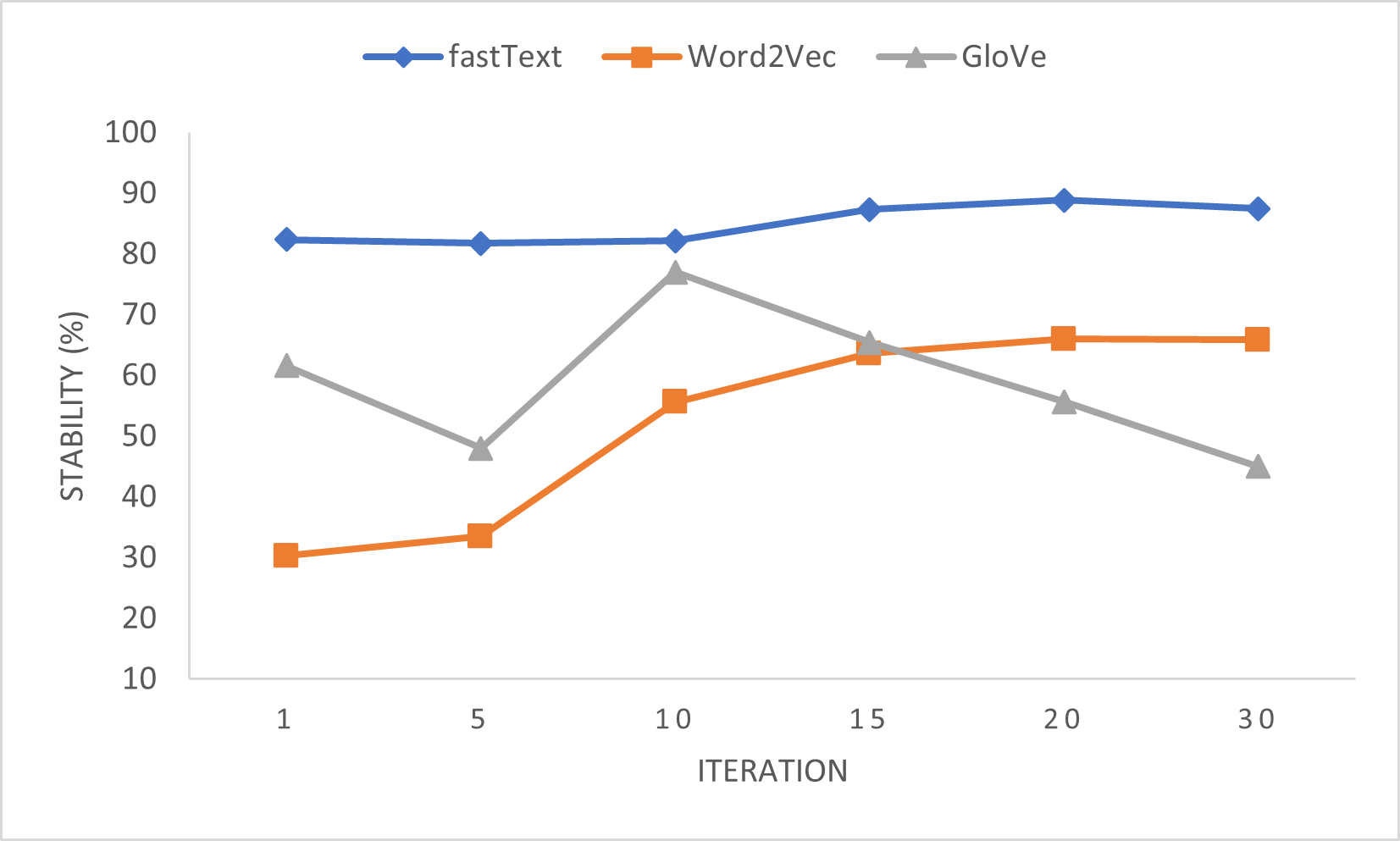}
         \caption{Europarl}
         \label{fig:ab}
     \end{subfigure}
     \hfill
     \begin{subfigure}[b]{0.4\textwidth}
         \centering
         \includegraphics[width=\textwidth]{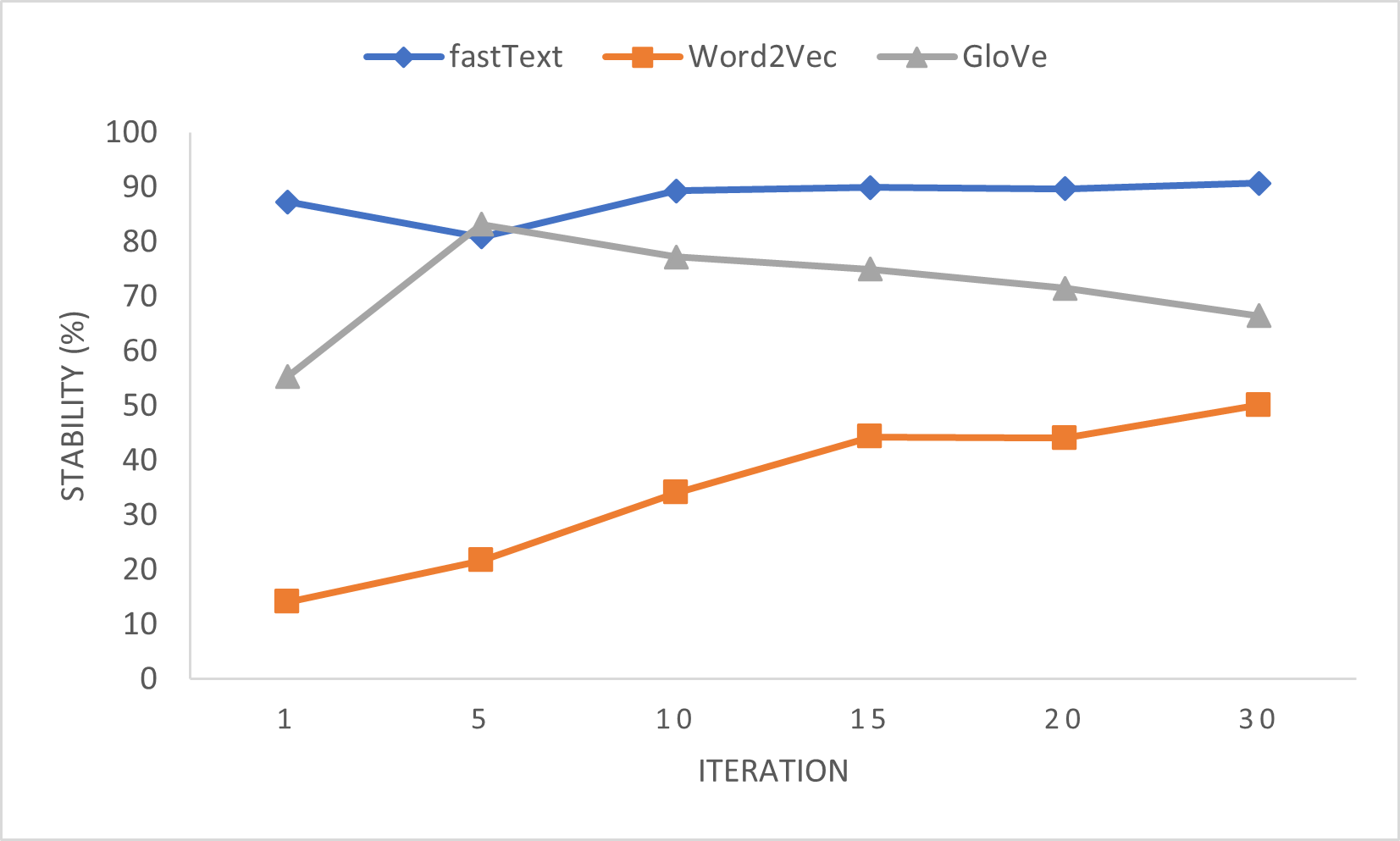}
         \caption{Lyrics}
         \label{fig:r}
     \end{subfigure}
     \hfill
     \begin{subfigure}[b]{0.4\textwidth}
         \centering
         \includegraphics[width=\textwidth]{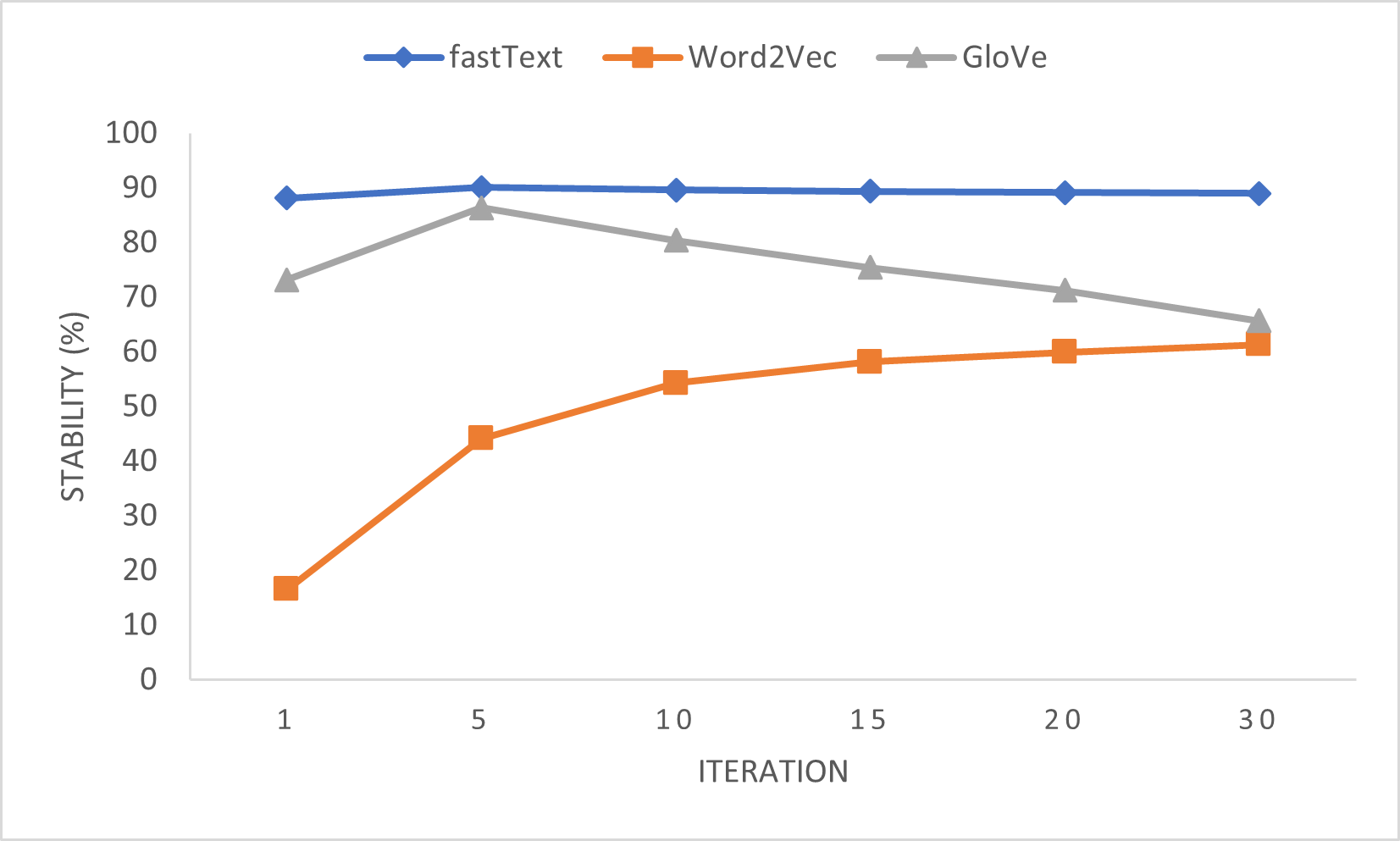}
         \caption{News 2007}
         \label{fig:s}
     \end{subfigure}
        \caption{Stability of WEMs with respect to the number of training epochs}
        \label{iter}
\end{figure}

While training the word embeddings, another design parameter is the number of training epochs. Ideally, we will expect the word embeddings to stabilize with increase in the number of training epochs. Figure \ref{iter} shows trends in the stability with variation in the number of training epochs. Given a particular number of training epochs on the X-axis, we are computing stability between two embeddings spaces that are trained for the same number of epochs. However, these embedding spaces differ in the random initial seeds. We can observe that variations in the stability of fasText are minimal. For Word2Vec, the stability improves significantly with increase in the training epochs. GloVe shows a rather interesting trend. The stability sharply increases as we increase the number of training epochs from 1 to 5. After that, stability starts falling significantly with further increase in the number of training epochs. This appears to contradict the intuition that increasing the number of training epochs will increase the stability. Furthermore, fastText performs better than both GloVe and Word2Vec for all the datasets. Stability of Word2Vec is inferior compared to the other two WEMs. However, its stability approaches that of GloVe with increase in the training epochs. Even after 30 training epochs, stability of Word2Vc and GloVe is below 60\%. This indicates that choice of initial seed matters for these two WEMs. In contrast, stability of fastText hovers around 90\%, indicating that choice of initial seed is insignificant.

\begin{figure}
     \centering
     \begin{subfigure}[b]{0.4\textwidth}
         \centering
         \includegraphics[width=\textwidth]{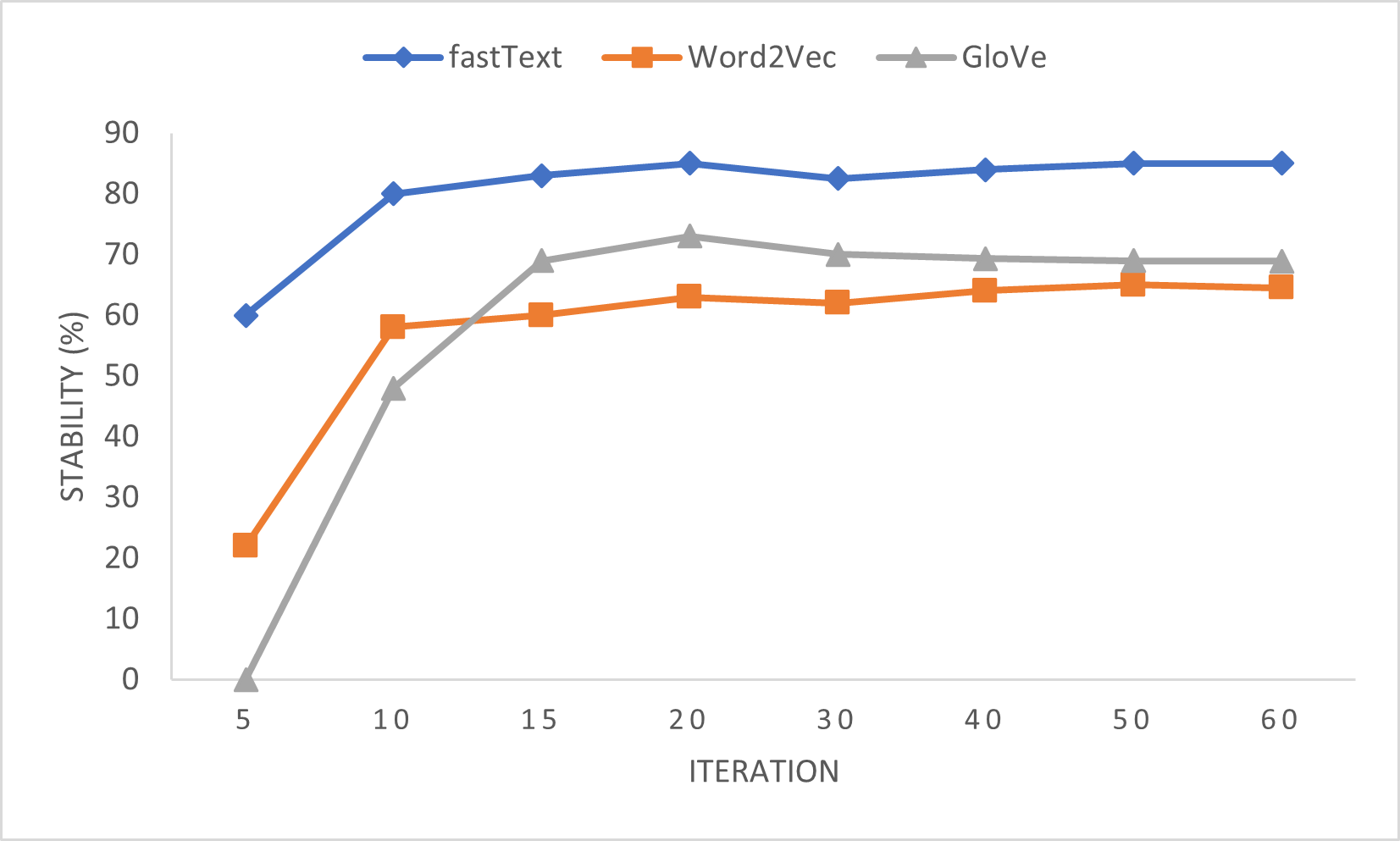}
         \caption{Wikipedia}
         \label{fig:4}
     \end{subfigure}
     \hfill
     \begin{subfigure}[b]{0.4\textwidth}
         \centering
         \includegraphics[width=\textwidth]{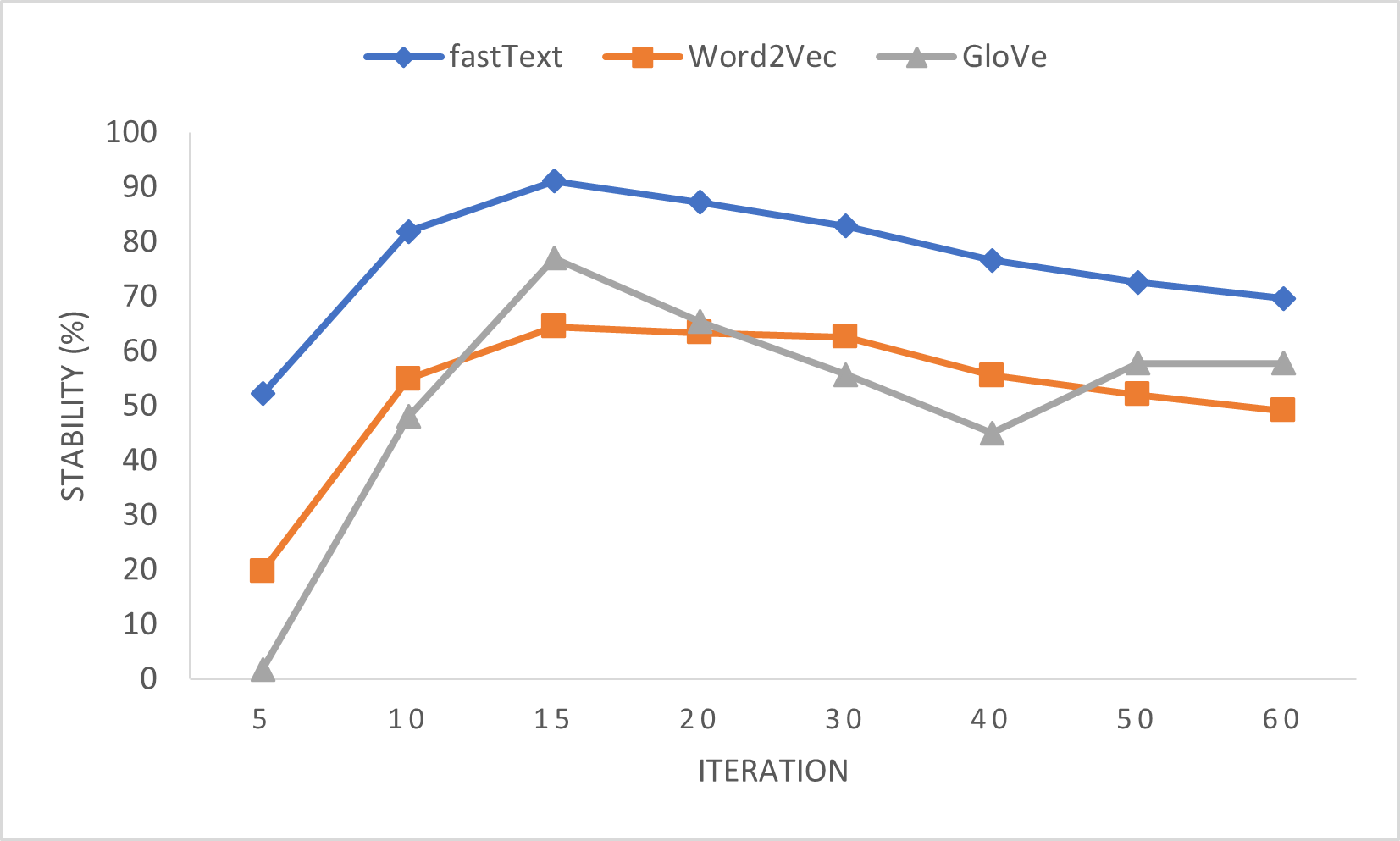}
         \caption{Europarl}
         \label{fig:ac}
     \end{subfigure}
     \hfill
     \begin{subfigure}[b]{0.4\textwidth}
         \centering
         \includegraphics[width=\textwidth]{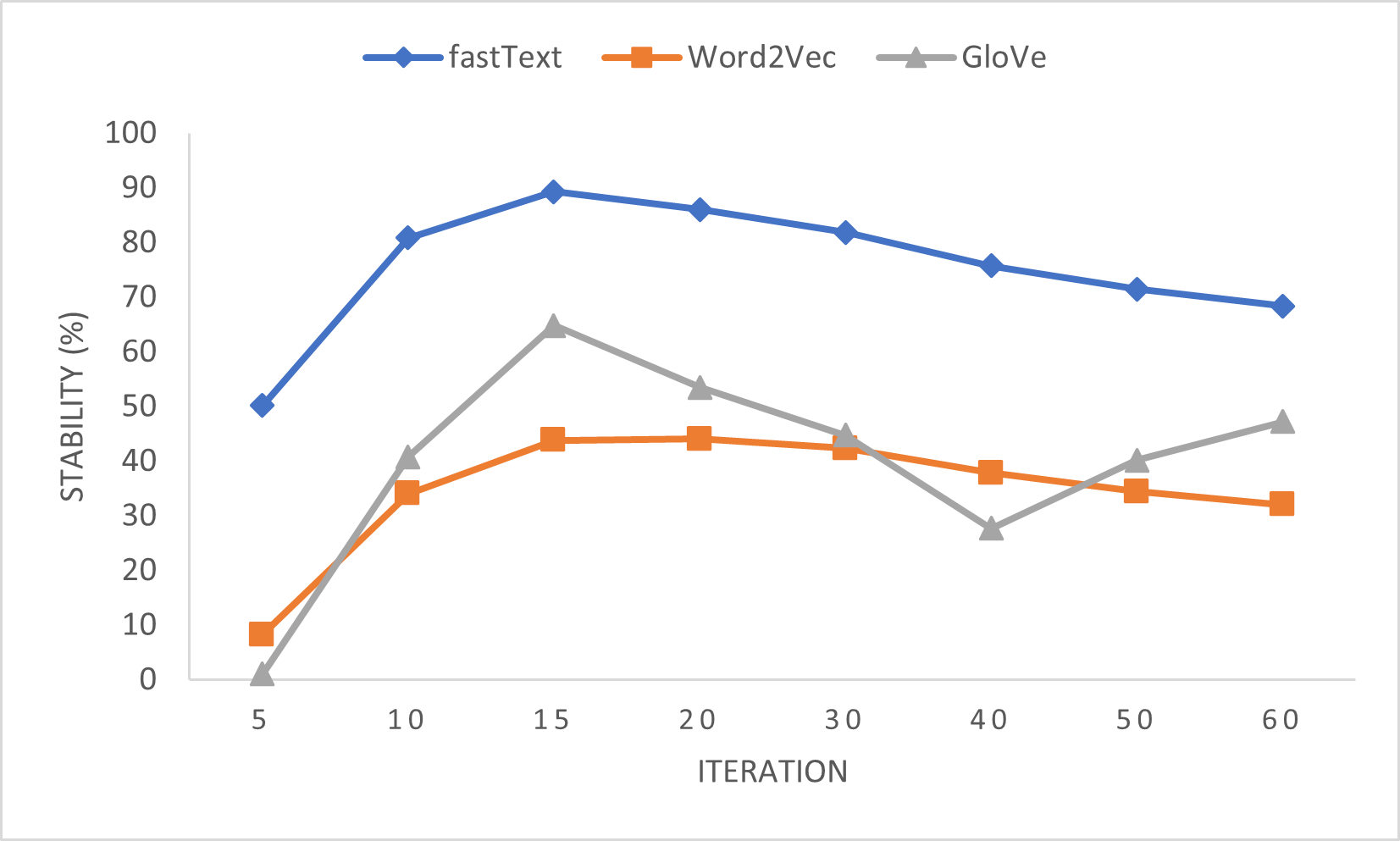}
         \caption{Lyrics}
         \label{fig:five over x}
     \end{subfigure}
     \hfill
     \begin{subfigure}[b]{0.4\textwidth}
         \centering
         \includegraphics[width=\textwidth]{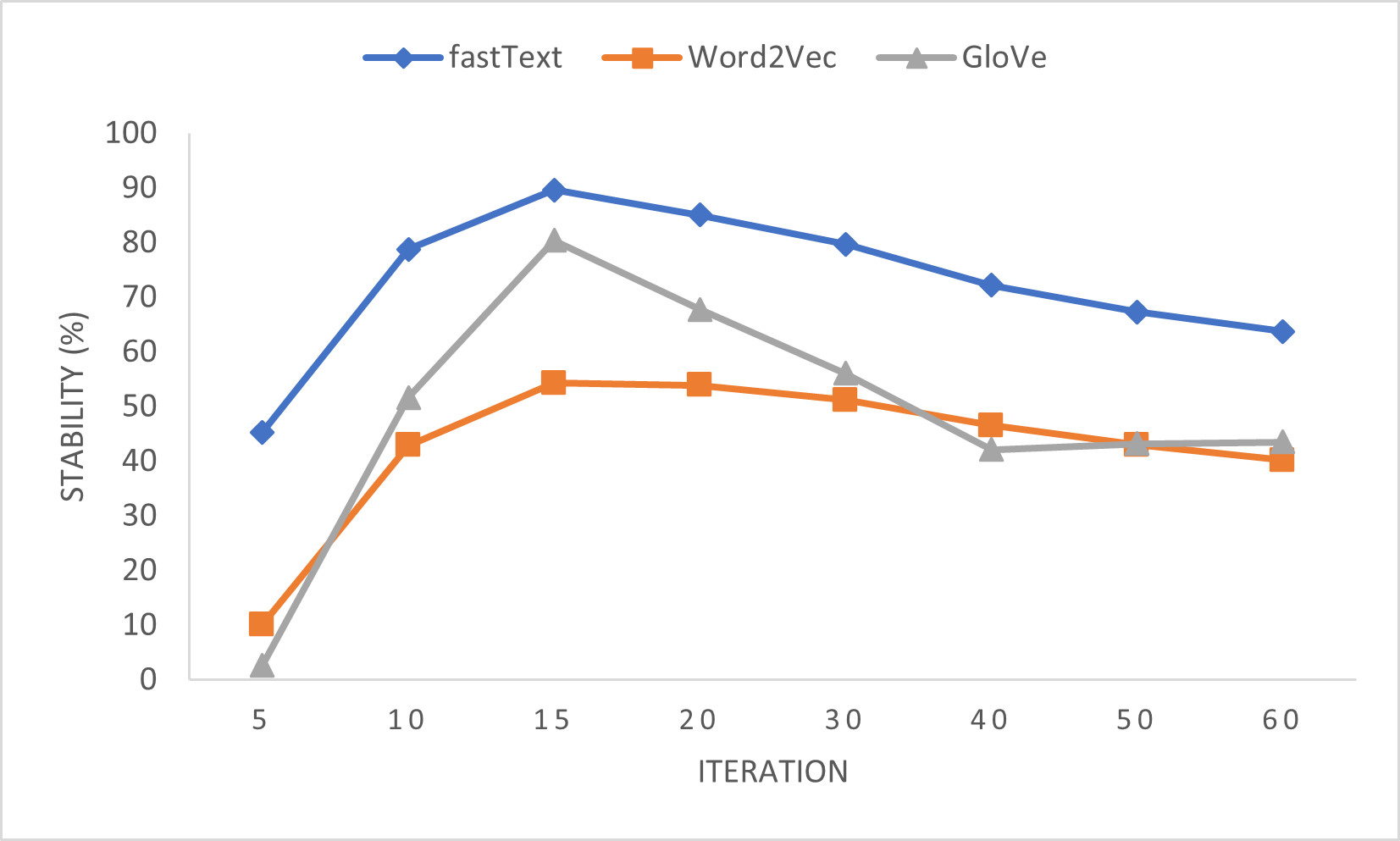}
         \caption{News 2007}
         \label{fig:y}
     \end{subfigure}
        \caption{Stability of WEMs across successive training epochs}
        \label{euro_new}
\end{figure}

We also wanted to quantify the changes in word embeddings across successive epochs. Please refer to Figure \ref{euro_new}. For a given WEM, we are comparing two embedding spaces $E_1$ amd $E_2$. Both these embedding spaces are identical to each other except for the number of training epochs. Let $Train(E_1)$ and $Train(E_2)$ be the number of training epochs for respective embedding spaces. X-axis in the Figure \ref{euro_new} represents $train(E_1)$. We set $train(E_2)=Train(E_1)-1$.  For example, using fastText as the WEM when we compare $E_1$ and $E_2$ with $Train(E_1)=5$ and $Train(E_2)=4$, the stability is close to 60\%. Even in this scenario, fastText is the most stable WEM. It indicates that word embeddings in fastText do not change significantly across successive iterations. So we can terminate the training of fastText after a few training epochs. In contrast, Word2Vec and GloVe have significantly lower stability across successive iterations. The stability value for this experiment does not reach close to 100\% for any of the WEMs. This indicates that  even with high number of training epochs, top ten nearest neighbors for many words do not stabilize.

\subsection{Effect of Context Window Size on Stability}
\begin{figure}
     \centering
     \begin{subfigure}[b]{0.4\textwidth}
         \centering
         \includegraphics[width=\textwidth]{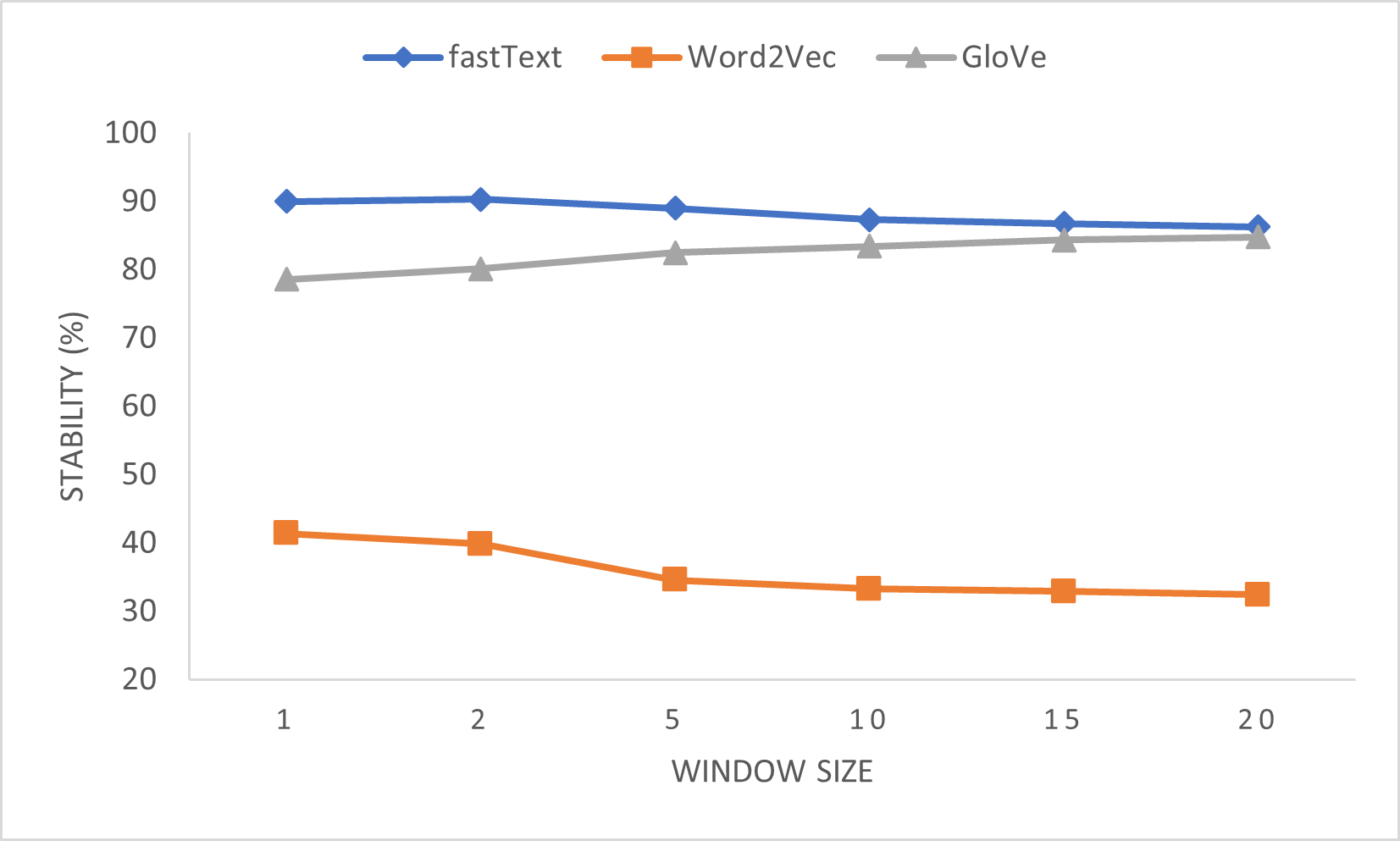}
         \caption{Wikipedia}
         \label{fig:5}
     \end{subfigure}
     \hfill
     \begin{subfigure}[b]{0.4\textwidth}
         \centering
         \includegraphics[width=\textwidth]{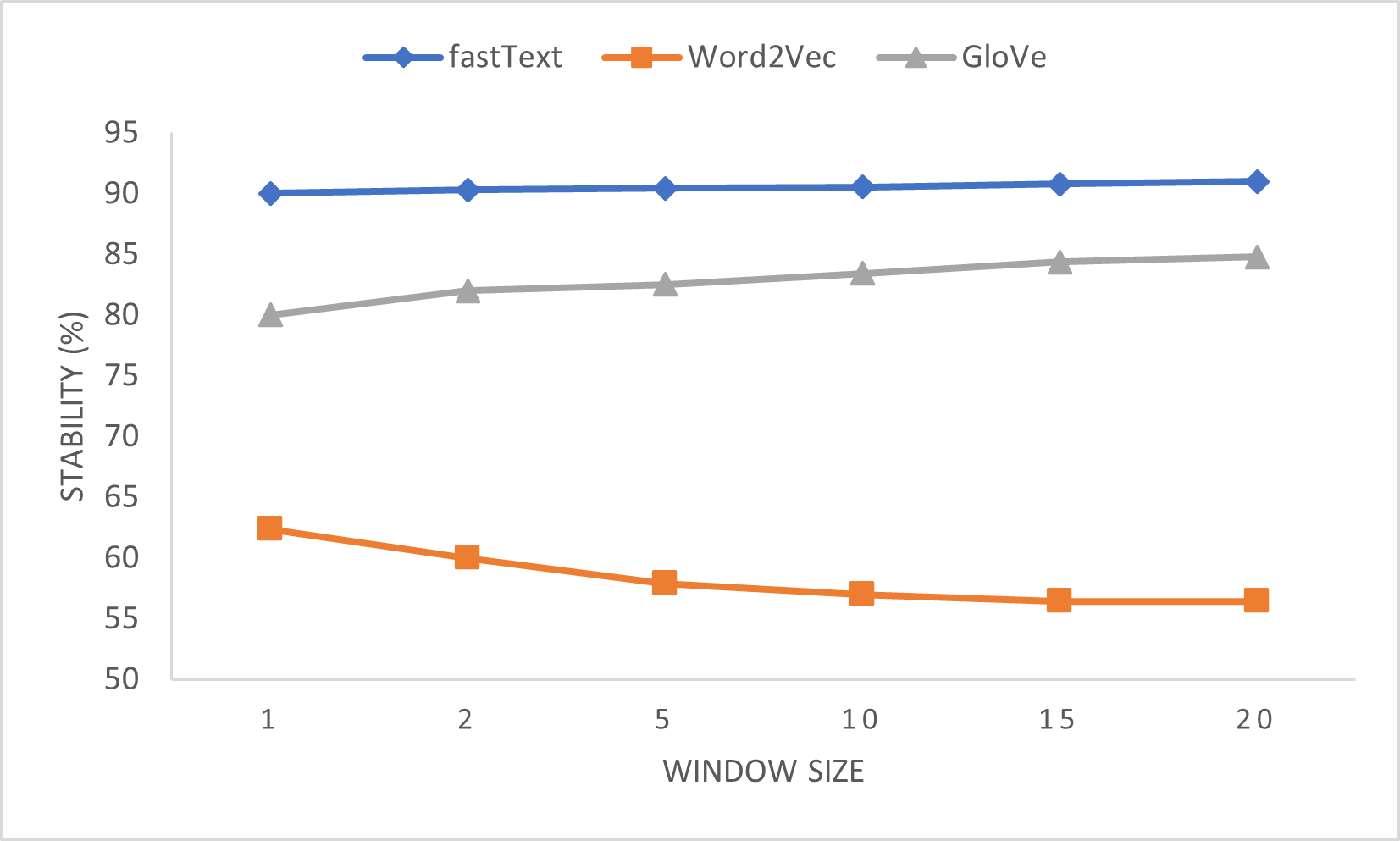}
         \caption{Europarl}
         \label{fig:ad}
     \end{subfigure}
     \hfill
     \begin{subfigure}[b]{0.4\textwidth}
         \centering
         \includegraphics[width=\textwidth]{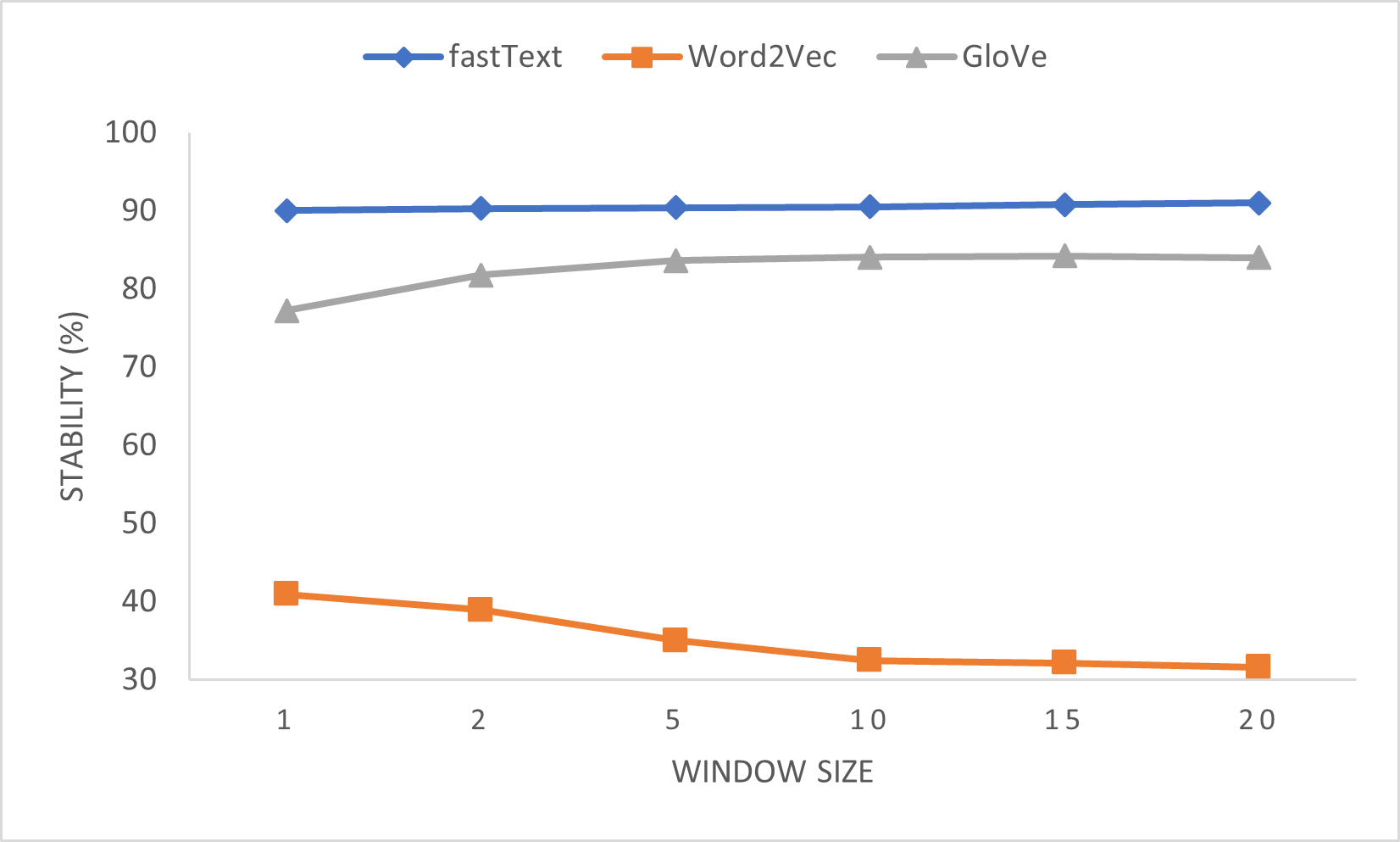}
         \caption{Lyrics}
         \label{fig:z}
     \end{subfigure}
     \hfill
     \begin{subfigure}[b]{0.4\textwidth}
         \centering
         \includegraphics[width=\textwidth]{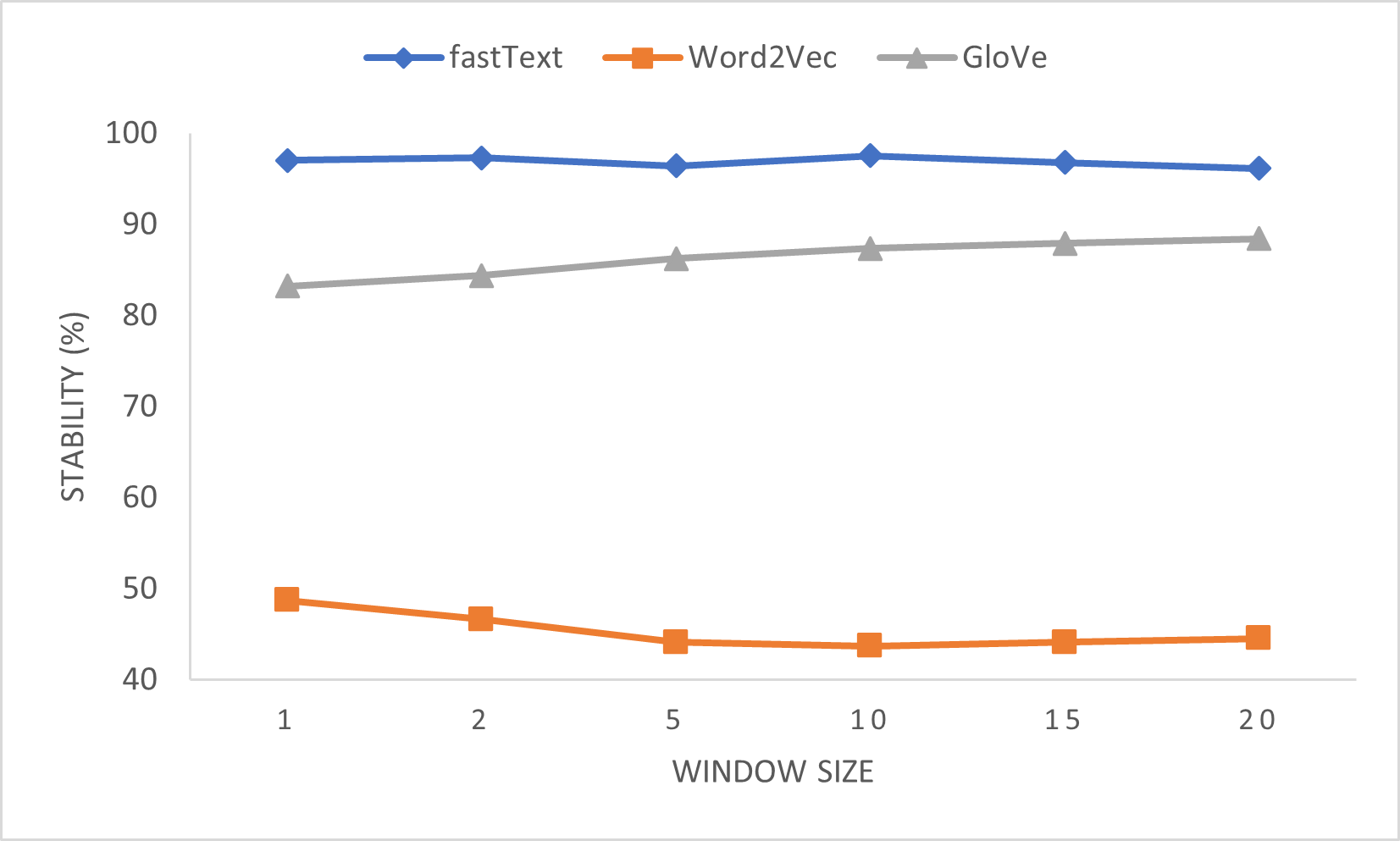}
         \caption{News 2007}
         \label{fig:a}
     \end{subfigure}
        \caption{Stability of WEMs with respect to the context window size}
        \label{wikiWindow}
\end{figure}

The intuition behind any WEM is to generate the representation for a word by learning from its context. The context is limited only to the sentence that the word belongs to. Size of the context window is also a design parameter while training the word embeddings. Figure \ref{wikiWindow} shows variation in WEM stability with respect to the context window size. For fastText, there is slight reduction in stability. Word2Vec and GloVe show contrasting behavior.After the context window size of ten, these is not significant change in the stability value for any WEM. This is expected as long sentences are rare in real-world datasets that we have considered. Even for this experiment, fastText turns out to be the most stable WEM.

\section{Effect of Stability on Downstream tasks}
In the previous section, we have analyzed various factors affecting the WEM stability. None of the WEMs are completely stable. Word embeddings are used for various downstream tasks. These embeddings typically serve as initial input for various task specific algorithms and machine learning models. The question we want to ask is: What is the effect of WEM instability on downstream tasks? For our experiments, we have shortlisted three tasks to cover multiple paradigms.  First is an unsupervised learning task of word clustering using SNND algorithm. Second is a supervised learning task of POS tagging using a Deep Learning model. Third is a fairness evaluation task involving subjective human opinions.

\begin{figure}[hbt]
    \includegraphics[width=8.4 cm,height=4.2cm]{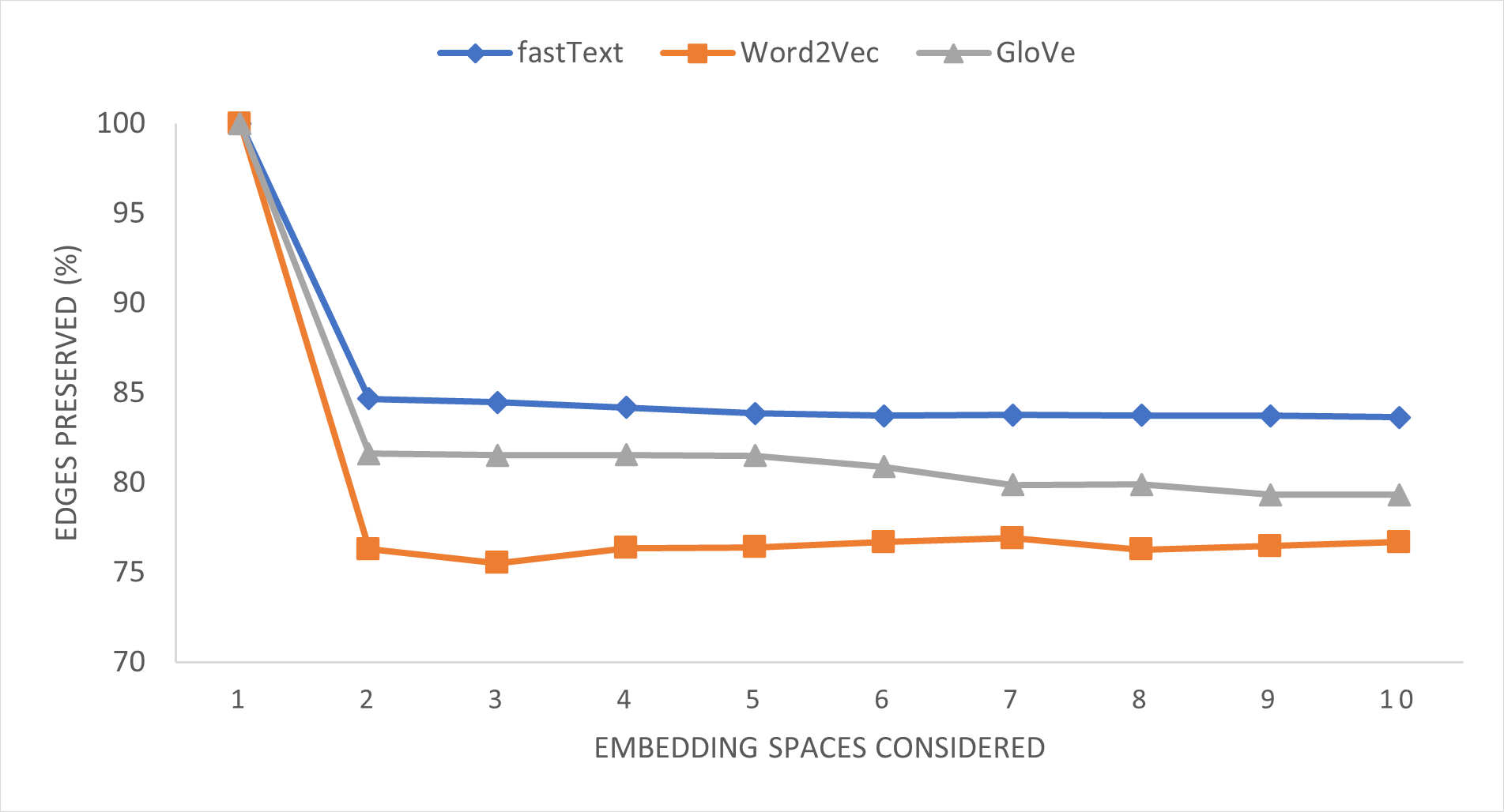}
    \caption{Decrease in clustering agreement plateaus with increase in the number of embedding spaces.}
    \label{cluster}
\end{figure}

\subsection{Word Clustering}
There is a vast variety of clustering algorithms available in the literature. Out of these algorithms we have chosen Shared Nearest Neighbor Density Based Clustering (SNND) algorithm \cite{doi:10.1137/1.9781611972733.5} for the following two reasons. First, SNND is a well known robust clustering algorithm for high-dimensional data. Second, this algorithm clusters data based on the shared top K-nearest neighbors (KNN). This intuition of clustering correlates with our stability measurement.

\begin{figure}[hbt]
    \includegraphics[width=8.4cm,height=4.2cm]{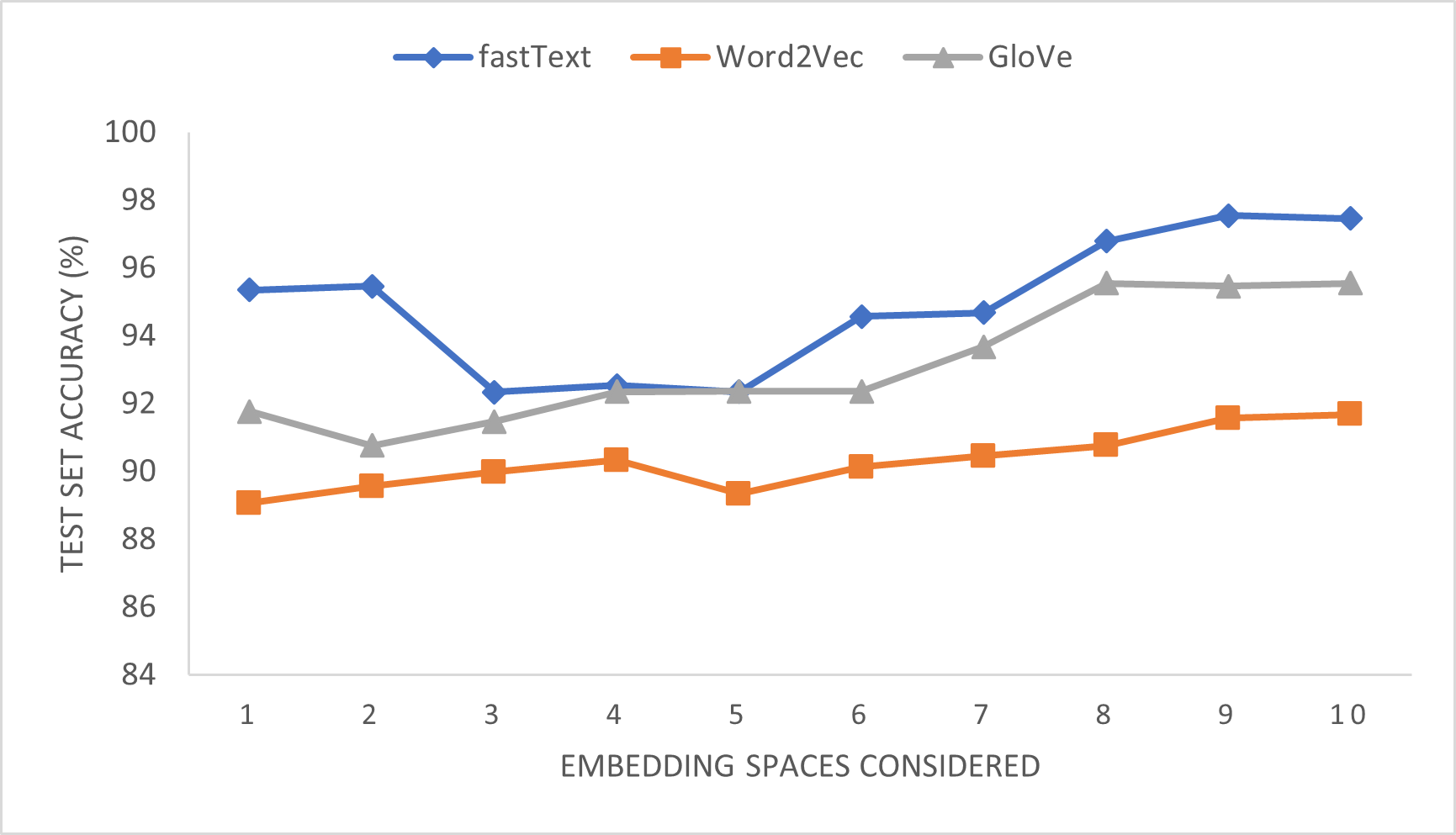}
    \caption{Accuracy of POS tagging task increases with inputs from multiple embedding spaces}
    \label{pos1}
\end{figure}

SNND is a combination of two influential clustering algorithms: Jarvis-Patrick \cite{Jarvis_Patrick} and DBScan \cite{10.5555/3001460.3001507}. In the first step, SNND builds a graph representation of data using the concepts from Jarvis-Patrick algorithm. Then it clusters this graph using concepts from DBScan algorithm. SNND begins by building KNN list for each data point. Similarity between any two data points is the extent of overlap between their KNN lists. Next step is to build the SNN-Graph. In this graph, each data point is a vertex and two vertices have an edge if their similarity is above a predefined threshold $\delta_{sim}$. A vertex is labelled as core if its degree is above another predefined threshold $\delta_{degree}$. A vertex is labelled as noise if it is not core and it is not connected to any other core point. The remaining vertices are labelled as non-core. All core vertices are initially considered as a separate clusters. In the merging phase, two clusters are merged if at least one core point from each cluster is connected with each other in the SNN-graph. This iterative step is repeated till no more clusters can be merged. Noise vertices are removed from clustering. For each non-core vertex, the cluster having the strongest link with it is chosen.

\begin{table*} [hbt]
\centering
\begin{tabular}{|p{0.5cm}|p{5cm}|p{3.5cm}|p{1.4cm}|p{1.2cm}|p{1.2cm}|p{1cm}|p{1 cm}|}
\hline
\textbf{Test  no.} & \textbf{Target Words} & \textbf{Attribute Words} & \textbf{fastText} & \textbf{GloVe} & \textbf{Word2Vec} & \textbf{CA}&\textbf{IAT}\\
\hline
    1 & flowers vs insects & pleasant vs unpleasant & 1.17,1.15 & 1.37,1.25 & 0.93,0.79 & 1.5 &1.35\\ \hline
    2 & instruments vs weapons & pleasant vs unpleasant & 1.64,1.65 & 1.49,1.45 & 1.68,1.62 &1.53& 1.66\\ \hline
    3 & european american vs african american names & pleasant vs unpleasant & 0.36,0.35 & 0.39,0.40 & 0.64,0.59 & 1.41&1.17 \\ \hline
    4 & male vs female names & pleasant vs unpleasant & 1.895,1.893 & 1.80,1.74 & 1.77,1.75 &1.81 &0.72 \\ \hline
    5 & maths vs arts & male vs female names & 0.86,0.84 & 1.10,1.16 & 0.49,0.25 & 1.06&0.82  \\ \hline
    6 & science vs arts & male vs female names & 1.09,1.08 & 1.28,1.22 & 0.42,0.29 &1.24& 1.47 \\ \hline
    7 & mental disease vs physical disease & temporary vs permanent & $-$0.94,$-$0.93 & 1.43,1.34 & 1.4,1.2 &1.38& 1.01\\  \hline
    8 & young people vs old people names & pleasant vs unpleasant & 0.83,0.80 & 1.16,0.87 & 0.06,$-$0.04 &1.21& 1.42 \\ \hline
\end{tabular}
\caption{\label{comparison-weat}
Stability of WEM correlates with the stability of WEAT results
}
\end{table*}

We experimented with all three WEMs on the Wikipedia (Sample) dataset to evaluate the effect of WEM instability on SNND clustering. Using each WEM, we train $P$ embedding spaces: $E_1$ to $E_{P}$. These embedding spaces differ only in the random initial seeds. Corresponding to each embedding space, we obtain the clusterings $C_1$ to $C_{P}$. We compute the agreement of these $P$ clusterings as the percentage of edges from $C_{1}$ that are preserved across all $P$ clusterings. Please refer to Figure~\ref{cluster}. This figure shows the variation in the clustering agreement while varying the value of $P$ from one to ten. As we increase the value of $P$, the clustering agreement decreases. However, this decrease plateaus for higher values of $P$. This indicates that even with the inherent instability of WEMs, resulting clustering always preserves significant majority of the edges. Here again, we observe that fastText results in the highest clustering agreement.

\subsection{Part-Of-Speech Tagging}

For POS tagging, we combine a number of corpora using the NLTK library. These include treebank (consists of 5\% of Penn Treebank), Brown, CONLL (Conference on Computational Natural Language Learning) 2000, 2002 and 2007. The total corpora size consists of 2.5 million tokens. We have split the data into train, validation and test sets in the ratio 70:15:15.

While performing the POS tagging task, our goal is not to replicate the state of the art. We want to analyze the effect of WEM instability on the task. Like the previous stability studies \cite{wendlandt2018factors}, we chose a simple a bidirectional LSTM model \cite{bidir}. It has 50 dimensional hidden vectors and the outputs are passed through a softmax layer.  Similar to the clustering task, we trained $P$ embedding spaces $E_1$ to $E_P$ using each of the WEMs. The model requires vector representation of words as input. While considering $P$ embedding spaces to generate the vector representation of a word, we simply compute the average of embeddings  across all $P$ embedding spaces. This averaging strategy across multiple embedding spaces is shown to be effective in the literature \cite{coates-bollegala-2018-frustratingly}. Please refer to the Figure \ref{pos1}. This figure shows variation in the POS tagging accuracy score with respect to the number of embedding spaces considered. Initially, the accuracy fluctuates with increase in the number of embedding spaces. However, after crossing seven embedding spaces, the accuracy increases and then it plateaus. Here the instability of WEMs actually benefits the task. If the WEMs were stable and provided similar input across multiple embedding spaces then we will not see any improvement in the accuracy score. However, the increase in the accuracy score indicates that diferent embedding spaces provide complementary information to the model.

\subsection{Fairness Evaluation Task}

Human actions reflect various biases in our thinking. For example, musical instruments might seem inspiring and weapons might seem harmful to some people. Implicit Association Test (IAT) is designed to measure such biases with human subjects \cite{iat}. The Word Embedding Association Test (WEAT) is proposed by Caliskan et al.\cite{caliskan}. The WEAT test is an improvisation over the IAT test such that biases can be measured directly from the word embeddings in the absence of humans subjects. This test computes the similarity between words using cosine similarity. Fairness evaluation task involves measuring biases in the the given set of word embeddings using the WEAT test. We selected this test for our experiments because its intuition of cosine based similarity correlates with our stability evaluation method.

The WEAT test takes in sets of target and attribute words. Target words refer to the set of words which denote a particular group based on a specific criterion such as instrument names (guitar, flute)and weapons (gun, sword). Attribute words refer to the set of words which denote the characteristics such as pleasant words (inspiring, creative) and unpleasant words (dangerous, harmful). Given two sets of target words and two sets of attribute words, the WEAT test checks if the null hypothesis holds. That is, if the word embeddings are unbiased then the  similarity of the two sets of target words with the attribute words should be almost same. If the result value for the WEAT test is close to zero then it indicates the absence of bias. Please refer to Table \ref{comparison-weat}. It lists eight target and attribute word set tests that we borrowed from Caliskan et al. \cite{caliskan}. The column $IAT$ reports the bias measurement results from Greenwal et al. using the human subjects \cite{iat}. The column $CA$ reports the results from Caliskan et al. They trained their word embeddings using a large scale crawl of the Web. 

\begin{table*}
\centering
\begin{tabular}{|p{1.8cm}|p{2.3cm}|p{4.8cm}|p{2cm}|p{5.8cm}|}
\hline
\textbf{Paper} & \textbf{Algorithms} & \textbf{Stability factors} & \textbf{Datasets (number of tokens)} & \textbf{Results}\\
\hline

    Levy et al.\cite{levy2015improving}, 2015 
    & PPMI matrix, SVD factorization, SGNS, and GloVe
    & $\bullet$ Preprocessing parameters: Dynamic Window Size, Subsampling and Deleting Rare Words \newline                                                  $\bullet$ Association Metric parameters: Shifted PMI, Context Disribution Smoothing \newline                              
    $\bullet$ Postprocessing parameters: Addition of context vectors, Eigenvalue weighting, Vector normalization  
    & Wikipedia - Aug 2013 dump (1.5B)
    & $\bullet$ No global advantage to any single approach \newline                 $\bullet$ Careful hyperparameter tuning can outweigh the importance of         adding more data \newline                                                    $\bullet$ SGNS outperforms GloVe in every task  
      $\bullet$ For both SGNS and GloVe, adding context vectors variant can result in substantial gains
    \\ \hline

    Hellrich and Hahn\cite{hellrich2016bad}, 2016
    & SGNS and SGHS
    & Neighborhood size, Word Frequency, Word Ambiguity and Number of Training Epochs
    & Google N-gram corpus: English Fiction (4.8B) and German (0.7B)
    & $\bullet$ Both accuracy and reliability are higher for SGNS than for SGHS for all tested combinations of languages and time spans, if 10 training epochs are used \newline
    $\bullet$ If only one training epoch is used: little difference in accuracy between SGNS and SGHS, but SGHS is clearly better in terms of reliability \newline
    \\ \hline
    
    Wendlandt et al.\cite{wendlandt2018factors}, 2018 
    & Word2Vec, GloVe and PPMI 
    & $\bullet$ Word properties: Primary and Secondary POS, no of syllables in a word, no of different POS present \newline
    $\bullet$ Training Data properties: Raw frequency in embedding spaces and their difference, vocabulary size of embedding spaces and their difference, percentage overlap between embedding spaces, curriculum learning, training position in embedding spaces and their difference \newline
    $\bullet$ Algorithm properties: embedding dimensions of embedding spaces and their difference
    & NYT (58M) and Europarl (61M)
    & $\bullet$ Curriculum learning is important \newline
      $\bullet$ POS highly affects stability \newline
      $\bullet$ GloVe is the most stable algorithm overall \newline
      $\bullet$ Frequency is not a major factor for stability \\ \hline
      
      Chugh et al.\cite{chugh} , 2018
      & Word2Vec
      & Embedding dimension and Frequency
      & Brown (10k), Project Gutenberg (10k) and Reuters (10k)
      & $\bullet$ Stability of high-frequency words increases and then stabilizes to a lower value. Stability of mid- and low-frequency words initially declines and increases to a stable value \newline
      $\bullet$ Embedding size is a causal factor and varies across corpora \\ \hline
      
      Antoniak and Mimno\cite{antoniak2018evaluating}, 2018
      & LSA, SGNS, GloVe, PPMI 
      & Corpus parameters: Order and presence of documents, Size of corpus and length of documents 
      & U.S. Federal Courts of Appeals (38k), NYT (22k) and Reddit (26k)
      & $\bullet$ Nearest-neighbor distances are highly sensitive
        to small changes in the training corpus \newline
        $\bullet$ All techniques are not sensitive to document order \newline   
        $\bullet$ All techniques are sensitive to the presence of specific documents \newline  
        $\bullet$ Variability initially increases with the number of models trained \\ \hline

     Dridi et al.\cite{dridi2018k}, 2018
     & Word2Vec (Skipgram)
     & Vector dimension, Window context and Vocabulary size
     & NIPS (Neural Information Processing System) between 2007 and 2012 (2M)
     & $\bullet$ Accuracy of Word2Vec increases with increasing corpora size \newline
     $\bullet$ Stability increases considerably as the dimensionality increases, but after reaching some point, it diminishes or becomes slightly invariant 
     \\ \hline

\hline
\end{tabular}
\caption{\label{related-works}
Summary of related work
}
\end{table*}

\balance
We ran the WEAT test on the three WEMs and used the Wikipedia (Sample) dataset for experimentation. For each WEM, we trained three sets of word embeddings. Using each set of word embeddigs, we ran the WEAT test. In Table \ref{comparison-weat}, three columns for the WEMs report stability results for each method. These three columns contain two values each. These two values correspond to the maximum and minimum value obtained for a test using the given WEM across multiple embedding spaces. For example, consider the WEAT results for test number 1 (flowers vs insects) using the fastText WEM. Corresponding to three embedding spaces, we obtained three different result values. In the table, we list the  maximum and minimum of these three values: $1.17$ and $1.15$. Ideally, our results should be close to the columns $CA$ and $IAT$. However, we have trained our word embeddings on much smaller dataset as compared to the dataset used by Caliskan et al. That is the reason why quality of our results does not match well with $CA$ and $IAT$. But our goal is not to demonstrate the quality of output. We aim to measure the stability of the results and compare it with the stability of the WEMs used. Till now, we have observed that fastText is the most stable WEM amongst the three methods. As a consequence, WEAT results for fastText have the highest stability. There is very little difference in the maximum and minimum values in the $FT$ column. WEAT results for Word2Vec are the least stable. This is expected as we have observed that Word2Vec is the least stable method among the three methods considered in this paper.

\section{Related Work}
There has been a growing interest among researchers to explore the stability of the word embedding techniques. Wendlandt and others \cite{wendlandt2018factors} showed how various factors contribute to the stability of word embeddings including frequency of words. They also analyzed the effects of stability on downstream tasks. Chugh and others \cite{chugh} presented the impact of dimension size and frequency of words on the consistency of a word embedding model using Word2Vec. Pierrejean and Tanguy \cite{pierrejean2018towards} explored the variations between different word embeddings models trained using Word2Vec. Their study shows that variation is influenced by parameters of a training algorithm. These parameters influence the geometry of word vectors and their context vectors \cite{mimno2017strange}, thereby causing variations. Hellrich and Hahn \cite{hellrich2016bad} assess the reliability of word embeddings for both modern and historical English and German and provided insights about optimal parameters for negative sampling method of Word2Vec and the number of iterations to train for. They also showed that negative
sampling outperforms hierarchical softmax in terms of reliability \cite{hellrich2016assessment}. Furthermore, stability is also affected by document properties as shown by Antoniak and Mimno \cite{antoniak2018evaluating}. Their study shows that smaller corpora are much more affected due to these effects. Related work is summarized in Table \ref{related-works}.

\section{Conclusion and Future Work}
In this work, we have shown that WEMs are not completely stable. Out of three three WEMs studied, fastText is the most stable and Word2Vec is the least stable method. We also observed the effect of WEMs instability on three downstream tasks. While training word embeddings, one needs to carefully analyze the effect of various design choices on the stability of word embeddings. In future, we need a deeper analysis and theoretical explanation for various stability trends observed in this paper.

\bibliographystyle{ACM-Reference-Format}
\bibliography{sample-base.bib}

\end{document}